\pdfoutput=1
\documentclass[verbose=true,letterpaper,margin=1in]{article}

\usepackage[main, final]{neurips_2026}

\usepackage[utf8]{inputenc} %
\usepackage[T1]{fontenc}    %
\usepackage{hyperref}       %
\usepackage{url}            %
\usepackage{booktabs}       %
\usepackage{amsfonts}       %
\usepackage{nicefrac}       %
\usepackage{microtype}      %
\usepackage{xcolor}         %

\usepackage{booktabs}
\usepackage{multirow}
\usepackage{makecell}
\usepackage{xcolor}
\usepackage{colortbl}
\usepackage{amsmath}
\usepackage{graphicx}
\usepackage{subcaption}
\usepackage[export]{adjustbox}
\usepackage{pifont}
\usepackage{enumitem}
\usepackage[capitalize]{cleveref}
\usepackage{comfortaa}       %
\usepackage{etoc}            %
\crefname{section}{Sec.}{Secs.}
\Crefname{section}{Sec.}{Secs.}
\crefname{subsection}{Sec.}{Secs.}
\Crefname{subsection}{Sec.}{Secs.}
\crefname{subsubsection}{Sec.}{Secs.}
\Crefname{subsubsection}{Sec.}{Secs.}
\usepackage{changepage} %

\definecolor{bestcolor}{RGB}{0, 100, 0}
\definecolor{secondcolor}{RGB}{0, 0, 180}
\definecolor{grayrow}{gray}{0.93}
\definecolor{upperbound}{gray}{0.85}
\definecolor{deltapos}{RGB}{28,115,60}
\definecolor{deltaneg}{RGB}{178,48,40}

\newcommand{\best}[1]{\textbf{#1}}
\newcommand{\second}[1]{\underline{#1}}

\definecolor{oursblue}{RGB}{224,238,255}
\definecolor{oursblue1}{RGB}{240,253,255}
\newcommand{\mnametitle}[1]{\textsc{OnlyAct}}
\newcommand{\mname}[1]{OnlyAct}

\definecolor{titlecolor}{RGB}{30,90,180}
\definecolor{linkpink}{RGB}{219,39,119}
\title{\textcolor{titlecolor}{\mnametitle{}}: Extracting Action Representations from Foundation Models for World Models}

\author{%
  Runjia Li \quad Minghao Chen \quad Junyu Xie \quad Philip Torr \quad Andrea Vedaldi \quad Tomas Jakab \\
  University of Oxford \\
  {\hypersetup{pdfborder={0 0 0}}\href{https://runjiali-rl.github.io/onlyact/}{\texttt{\textcolor{linkpink}{runjiali-rl.github.io/onlyact/}}}} \\
}

\hypersetup{pdftitle={OnlyAct: Extracting Action Representations from Foundation Models for World Models}, pdfauthor={Runjia Li, Minghao Chen, Junyu Xie, Philip Torr, Andrea Vedaldi, Tomas Jakab}}

\begin{document}

\maketitle
\begin{figure*}[h]
    \centering
    \includegraphics[width=0.98\textwidth]{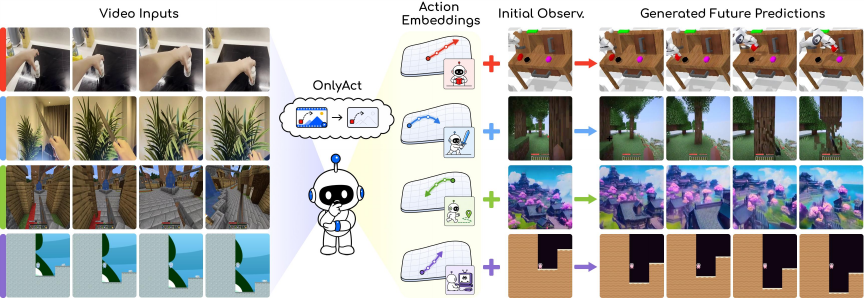}
    \vspace{-1mm}
    \caption{\textbf{\mname{}: an unlabeled action representation learning framework} builds on a pretrained foundation model to extract \textit{expressive} and \textit{transferable} action representations, keeping only the action while discarding the scene in which it occurs. Actions from diverse domains are mapped into a single shared action space, inheriting broad spatio-temporal priors from large-scale video pretraining. These action embeddings can be derived from video frames alone and used to generate videos with the same actions from different initial images across diverse domains. Real actions can also be adaptively mapped into the same action space for real-world control.}
    \label{fig:teaser}
    \vspace{-3mm}
\end{figure*}
\bigbreak

\definecolor{darkorange}{rgb}{1.0, 0.55, 0.0}
\newcommand{\editor}[1]{\textcolor{red}{[##1]}}
\newcommand{\alex}[1]{\textcolor{blue}{[\textbf{Alex} #1]}}
\newcommand{\willi}[1]{\textcolor{red}{[\textbf{Willi} #1]}}
\newcommand{\moayed}[1]{\textcolor{green}{[\textbf{Moayed} #1]}}
\newcommand{\sergey}[1]{{\color{cyan}{[\textbf{Sergey} #1]}}}
\newcommand{\arpit}[1]{{\color{green}{[\textbf{Arpit} #1]}}}
\newcommand{\runjia}[1]{{\color{magenta}{[\textbf{Runjia} #1]}}}
\newcommand{\ivan}[1]{{\color{orange}{[\textbf{Ivan} #1]}}}

\newcommand{\cellfirst}[1]{\textbf{#1}}
\newcommand{\cellsecond}[1]{\underline{#1}}
\newcommand{\cellthird}[1]{{#1}}

\newcommand{\cellred}{\cellcolor{Red!20}}
\newcommand{\cellorange}{\cellcolor{Orange!25}}
\newcommand{\cellyellow}{\cellcolor{Yellow!25}}
\newcommand{\cellgreendarkdark}{\cellcolor{Green!40}}
\newcommand{\cellgreendark}{\cellcolor{Green!30}}
\newcommand{\cellgreen}{\cellcolor{Green!20}}
\newcommand{\cellgreenlight}{\cellcolor{Green!10}}

\newcommand{\apref}[1]{Appx.~\ref*{#1}}
\newcommand{\website}{\emph{Website}}
\newcommand{\supp}{\emph{Appendix}}

\newcolumntype{C}[1]{>{\centering\arraybackslash}p{#1}}
\newcommand{\ourmodel}{}

\newcommand{\generator}{\mathcal{G}}
\newcommand{\patchinput}{\mathcal{P}_{I}}
\newcommand{\patchoutput}{\mathcal{P}_{O}}
\newcommand{\readlayer}{\mathcal{R}}
\newcommand{\writelayer}{\mathcal{W}}

\newcommand{\noise}{\boldsymbol{\epsilon}} %
\newcommand{\token}{\boldsymbol{x}}

\newcommand{\diffinput}{\mathbf{x}}

\newcommand{\inputtensor}{\mathbf{X}}
\newcommand{\inputtensorsource}{\mathbf{X}^{src}}
\newcommand{\inputtensortarget}{\mathbf{X}^{tgt}}
\newcommand{\inputtensortargetnoise}{\mathbf{X}^{tgt}_{0}}
\newcommand{\inputtensortargettone}{\mathbf{X}^{tgt}_{t^1}}
\newcommand{\inputtensortargetttwo}{\mathbf{X}^{tgt}_{t^2}}
\newcommand{\inputtensortargettthree}{\mathbf{X}^{tgt}_{t^3}}
\newcommand{\inputtensortargetclean}{\mathbf{X}^{tgt}_{1}}
\newcommand{\inputtensortargettime}{\mathbf{X}^{tgt}_{t}}
\newcommand{\inputtensortime}{\mathbf{X}_{t}}
\newcommand{\inputtensornoise}{\mathbf{X}_{0}}
\newcommand{\inputtensorclean}{\mathbf{X}_{1}}
\newcommand{\conditioning}{{c}}

\newcommand{\mask}{\mathbf{M}}

\newcommand{\velocity}{{v}}
\newcommand{\velocitypred}{{\hat{v}}}
\newcommand{\spatialact}{{s}}
\newcommand{\latentact}{{l}}
\newcommand{\spatialactca}{{s}_{CA}}
\newcommand{\latentactca}{{l}_{CA}}
\newcommand{\spatialactin}{{s}_{I}}
\newcommand{\latentactin}{{l}_{I}}
\newcommand{\spatialactout}{{s}_{O}}
\newcommand{\latentactout}{{l}_{O}}
\newcommand{\instruction}{y}

\newcommand{\hidsize}{{d}}
\newcommand{\numblocks}{{B}}
\newcommand{\numblocksstart}{{B}_\mathrm{in}}
\newcommand{\numblocksmid}{{B}_\mathrm{core}}
\newcommand{\numblocksend}{{B}_\mathrm{out}}

\newcommand{\numgroups}{{G}}
\newcommand{\latenttokcount}{{K}}
\newcommand{\latenttokcountgroup}{{J}}
\newcommand{\latenttokcountgroupmasked}{{\tilde{J}}}
\newcommand{\latenttokcountgroupmin}{{J_\mathrm{min}}}
\newcommand{\latenttokcountgroupmax}{{J_\mathrm{max}}}
\newcommand{\latenttokcountgroupuncond}{J_{\mathrm{uncond}}}
\newcommand{\latenttokcountgroupweak}{\tilde{J}_{\mathrm{w}}}
\newcommand{\latenttokcountgroupcond}{J_{\mathrm{cond}}}

\newcommand{\dataset}{\mathcal{D}}
\newcommand{\numtokens}{N}

\newcommand{\height}{H}
\newcommand{\width}{W}
\newcommand{\frames}{T}
\newcommand{\patchheight}{H_p}
\newcommand{\patchwidth}{W_p}
\newcommand{\patchframes}{T_p}

\newcommand{\tradeoffeff}{\rho}

\newcommand{\distributionscale}[1]{p^{{#1}}_t}

\newcommand{\datadistribution}{p_{d}}
\newcommand{\timestepdistribution}{p_{t}}
\newcommand{\noisedistribution}{p_{n}}
\newcommand{\difftimestep}{t}
\newcommand{\samplingthreshold}{\tau}
\newcommand{\sigmanoise}{\sigma_{n}}
\newcommand{\snr}{\mathit{SNR}}

\newcommand{\fidtenk}{\text{FID}_{10\text{K}}}
\newcommand{\fddtenk}{\text{FDD}_{10\text{K}}}
\newcommand{\fvdtenk}{\text{FVD}_{10\text{K}}}
\newcommand{\fidfifk}{\text{FID}_{50\text{K}}}
\newcommand{\fddfifk}{\text{FDD}_{50\text{K}}}

\definecolor{mediumtealblue}{rgb}{0.0, 0.33, 0.71}
\definecolor{darkpastelgreen}{rgb}{0.01, 0.75, 0.24}
\definecolor{azure}{rgb}{0.0, 0.5, 1.0}

\definecolor{crimsonred}{rgb}{0.86, 0.08, 0.24}
\definecolor{firebrick}{rgb}{0.7, 0.13, 0.13}
\definecolor{carmine}{rgb}{0.59, 0.0, 0.09}
\definecolor{rosewood}{rgb}{0.4, 0.0, 0.04}
\definecolor{deepcherry}{rgb}{0.6, 0.13, 0.13}

\begin{abstract}
Video world models promise general-purpose interactive simulators, but their training is limited by action supervision: action labels are scarce, fragmented across incompatible specifications, and do not scale with the unlabeled video pretraining corpus.
A common workaround jointly trains an action encoder to compress consecutive frames into a latent action and a decoder to reconstruct the next frame from the previous frame and that latent, forcing the latent to capture only new information.
This has two limitations: learning the action space from scratch can limit generalization to unseen domains, and the bottleneck trades expressiveness for separation, with tighter settings losing nuance and looser ones risking scene leakage under domain shifts. We argue for a different starting point: rather than learning an action representation from scratch, we start from a pretrained foundation model (V-JEPA~2), whose embeddings of a frame pair already carry substantial action information after large-scale, diverse video pretraining. This provides a more generalizable foundation for action representation. The remaining challenge is to extract the action while discarding the scene. To this end, we propose a label-free contrastive adaptation framework that pulls together frame pairs sharing the same motion under a different scene and pushes apart pairs from the same scene with different motion, reshaping the geometry of the pretrained space without imposing a reconstruction bottleneck.
We introduce a unified evaluation protocol that probes action representations along three axes---action expressiveness, scene invariance, and action-conditioned video generation quality---across robot manipulation, camera navigation, game, and driving benchmarks, most of which are unseen during adaptation.
Compared with prior annotation-free latent-action methods, our representations are substantially more expressive while matching their scene invariance, and they yield world models that transfer demonstrated actions to new scenes more faithfully.
\end{abstract}

\section{Introduction}

Over the past few years, video generative models~\cite{brooks2024sora, veo3_2025, HunyuanVideo, wan2025wan} have reached unprecedented levels of realism.
They are trained on vast quantities of video data and acquire broad priors over scene structure and dynamics, which has motivated their use well beyond video generation itself.
A growing line of work explores how they can be turned into video world models by conditioning generation on action signals~\cite{bruce2024genie, parker2024genie2, deepmind2025genie3, gao2025adaworld}.
Video world models have the potential to serve as general-purpose interactive simulators, with applications ranging from robotics to gaming.

Realizing this potential, however, hinges on action supervision.
While unlabeled video is abundant, only a small fraction is paired with the actions that produced it, and the labels that do exist are fragmented across incompatible specifications, such as joint angles in robotics or controller inputs in games.
Direct action conditioning therefore does not scale with the unlabeled video pretraining corpus and does not transfer well across environments.
To overcome this, latent-action models~\cite{menapace2021pvg, bruce2024genie} extract latent actions from unlabeled video in a self-supervised manner and use them to condition the video generator during pretraining.
Real actions can then be mapped onto this latent space afterward with a small amount of labeled data~\cite{bruce2024genie, gao2025adaworld}.

Latent-action models are typically trained from scratch as inverse-dynamics models: given a pair of consecutive frames, an action encoder infers a latent intended to capture the action, while a decoder reconstructs the next frame from the previous frame together with this latent. A bottleneck on the latent forces it to encode only what the decoder cannot recover from the previous frame alone, ideally isolating the action signal from scene content. This comes with two limitations: first, training the action space from scratch can prevent it from generalizing to unseen domains; second, tightening the bottleneck improves separation between action and scene at the cost of expressiveness, while loosening it preserves nuance at the cost of scene leakage, which can also hurt generalization when scene, embodiment, or dynamics shift.

In this work, we propose \mname{}, an unlabeled action representation learning framework that takes a different starting point.
Rather than learning the action representation from scratch, we start from a pretrained foundation model, V-JEPA~2~\cite{vjepa2}, and extract actions from it with contrastive learning.
Pretrained on large-scale, diverse video, such a model encodes both what is in a scene and how it changes over time, so the action-relevant signal is already present in its embeddings, alongside scene content.
We confirm this empirically: when a pair of consecutive frames is encoded as a two-frame clip, even the frozen V-JEPA~2 embedding carries more action information than latent-action models trained on the same data, as measured by a linear probe (\cref{tab:action_probing}). However, it also carries more scene information (\cref{tab:semantic_leakage}), and its probing performance is unstable across domains.
The main challenge is therefore not to learn an action representation from scratch, but to disentangle the action-relevant signal from scene content.

We do this by adapting the pretrained foundation model into an action encoder on unlabeled video with a contrastive objective and a lightweight low-rank adapter.
Given two frames from a video, we form a positive example that keeps the motion but changes the scene: when the camera is static, we use optical flow to cut out the moving region and paste it, at a shifted location, onto the background of a different scene. Hard negatives are drawn from other frame pairs of the same video, where action provides the dominant axis of variation. Together, these signals shift the encoder embedding space toward action.
Crucially, this is achieved without a reconstruction bottleneck: the objective reshapes the geometry of the pretrained space rather than compressing the action into a small code from which the next frame must be decoded.

The adapted encoder supports two main uses: transfer from demonstrations, and direct control with real actions. In the first, the encoder extracts a latent action from a pair of demonstration frames; this latent then conditions a video world model to reproduce the same action in a different scene.
In the second, a small action-labeled dataset can be used to map an environment's real actions onto the latent space, after which the video world model can be controlled directly through them.

Both uses require an action representation that captures rich and discriminative action dynamics, isolates the action signal from scene content so the same action transfers across scenes, and drives faithful action-conditioned video generation across different domains.
We construct a unified evaluation protocol around these axes and benchmark against prior annotation-free latent-action methods and the frozen pretrained encoder on thirteen robot, camera-navigation, and game datasets, ten of which are unseen during adaptation, as well as on world-model training and fine-tuning across robot, navigation, game, and driving domains.
Our representation is the most expressive on almost all datasets, matches the scene invariance of bottlenecked latent-action models, and yields world models with markedly better action transfer and controllability.

Our main contributions are as follows:
\begin{enumerate}[leftmargin=2.6em, itemindent=1pt, itemsep=1pt, parsep=2pt, topsep=-2pt]
    \item We introduce a new perspective on unlabeled action representation learning: rather than learning actions from scratch, we extract them from the representation space of a pretrained foundation model.

    \item We propose a label-free contrastive adaptation framework, built on flow-guided scene-swapped positives and same-video hard negatives, that disentangles action from scene without a reconstruction bottleneck and improves both action expressiveness and scene invariance over the frozen encoder.

    \item We introduce a unified evaluation protocol that systematically probes action expressiveness, scene invariance, and action-conditioned video generation quality, and show across robot, navigation, game, and driving benchmarks that our representations improve over prior annotation-free latent-action baselines.
\end{enumerate}

\noindent Code and models will be released on our project page: {\hypersetup{pdfborder={0 0 0}}\href{https://runjiali-rl.github.io/onlyact/}{\texttt{\textcolor{linkpink}{runjiali-rl.github.io/onlyact/}}}}.

\section{Related Work}

\textbf{Video World Models.}
Action-conditioned video world models have become a core approach for modeling environment dynamics in domains such as gaming and robotics. Existing methods typically obtain action information in one of two ways. The first assumes explicit action or intervention supervision, such as controller inputs, robot states, camera trajectories, or action prompts~\cite{diamond, unisim, cosmos, gamengen, zhang2025matrixgame, he2025matrixgame2, 2026matrixgame3, huang2025voyager, parker2024genie2, deepmind2025genie3, mao2025yume, Li_2025_ICCV, lingbot-world, xiao2025worldmemlongtermconsistentworld, pondaven2026actionparty}. This provides a precise control interface, but the supervision is tied to particular environments, embodiments, or data-collection pipelines, and therefore does not naturally scale to uncurated in-the-wild video. The second introduces latent-action interfaces learned from video~\cite{bruce2024genie, gao2025adaworld, gao2026dreamdojo, lapa, genie_redux, univla}, reducing the need for explicit labels but requiring a separate action-learning process. Current interactive video world models therefore face a trade-off between precise but environment-specific action supervision and scalable but separately learned latent actions. Our work is complementary: we use pretrained foundation-model representations as an effective starting point and ask whether they already contain transferable transition structure that can be extracted for world model pretraining.

\textbf{Latent Action Models from Video.}
A growing line of work learns latent actions from unlabeled video by jointly training an action encoder on frame pairs together with a future-frame
decoder, with a bottleneck on the latent.
Introduced by earlier work~\cite{edwards2019ilpo, menapace2021pvg, menapace2022playable} and scaled by recent methods~\cite{bruce2024genie, schmidt2024lapo, lapa, gao2025adaworld, gao2026dreamdojo, genie_redux, univla}, these latents serve as proxy actions for controllable video generation or policy learning.
The key limitation is generalization: because these encoders are trained from scratch with reconstruction objectives, the learned action space is shaped by the scenes, embodiments, and dynamics present in the training data.
As a result, latents can encode scene-specific cues such as appearance, object identity, or viewpoint rather than the action itself, making transfer difficult when scene or embodiment shifts~\cite{dai2026conla, zhang2026clap}.
Prior methods reduce this leakage with low-dimensional bottlenecks, discrete codebooks, or other architectural constraints, but these constraints come at the cost of reducing the resolution of the action space.
A separate line of work augments reconstruction-based latent-action models with contrastive objectives, but relies on additional supervision such as action labels or kinematic trajectories~\cite{dai2026conla, zhang2026clap}.
We instead start from a pretrained foundation model~\cite{vjepa2} and use a contrastive objective to extract action as a signal already present in its embeddings, neither learning the action space from scratch nor requiring additional supervision.

\textbf{Pretrained Representations.}
Pretrained visual representations~\cite{r3m, vip, mvp, vc1, voltron, dino-wm, karypidis2025dinoforesight, boduljak2025vfmf} provide strong and transferable features for downstream perception, control, and world modeling.
Large-scale foundation models such as V-JEPA~2~\cite{vjepa2} extend this to predictive spatio-temporal representations that support video understanding, and V-JEPA~2-AC further post-trains an action-conditioned predictor in this representation space for planning with robot actions.
These works use pretrained representations to encode observations or predict future states, with actions, when needed, supplied as labels.
We instead use the pretrained foundation model itself as an \emph{action} encoder: encoding a frame pair as a short clip exposes the motion between the frames, and we adapt the encoder without labels so that this motion signal, rather than scene content, dominates its embedding space.

\section{Method}
\label{sec:method}
\begin{figure*}[t]
    \centering
    \includegraphics[width=\textwidth]{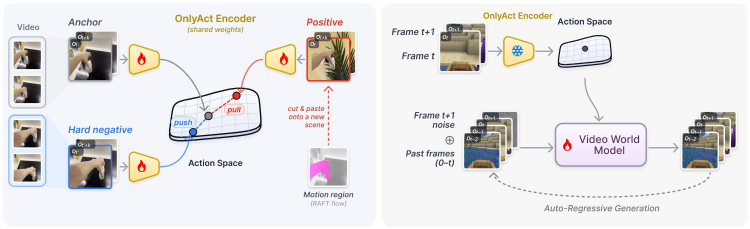}
    \caption{
    \textbf{Overview of the proposed learning pipeline.}
    \textbf{Left: Contrastive action adaptation.}
    An anchor frame pair $(o_t, o_{t+k})$ is encoded as a two-frame clip by the \mname{} encoder $\phi_\theta$, a pretrained foundation model with low-rank adapters whose weights are shared across all branches.
    The positive keeps the action but changes the scene: the moving region of the anchor, found by optical flow, is cut out and pasted at a shifted location onto the background of a different scene (for moving cameras, a different photometric augmentation of the anchor is used instead).
    The hard negative is drawn from a non-overlapping window $(o_{t'}, o_{t'+k})$ of the same video, sharing the scene but corresponding to a different action.
    In the action space, the anchor is pulled towards its positive and pushed away from its hard negative.
    \textbf{Right: Action-conditioned video world model.}
    The frozen \mname{} encoder $\phi_\theta$ maps consecutive frame pairs into the learned action space, yielding an action stream $\alpha_{1:t}$ that conditions an autoregressive video world model $G_\psi$.
    Given past frames $o_{1:t}$ and the action stream, $G_\psi$ predicts the next frame $o_{t+1}$, and is rolled out autoregressively to generate subsequent frames.}
    \label{fig:method_overview}
\end{figure*}

Given a corpus of videos without action labels, each consisting of frames $(o_1, \dots, o_T)$, our goal is to learn a latent action $\alpha_t$ for each consecutive frame pair $(o_t, o_{t+1})$ and use these latent actions to pretrain an action-conditioned video world model $G_\psi$ that predicts the next frame $o_{t+1}$ from $(o_{1:t}, \alpha_{1:t})$.
Our pipeline has two stages: we first adapt a pretrained foundation model $\phi_\theta$ on unlabeled video so that $\alpha_t = \phi_\theta(o_t, o_{t+1})$ isolates the action signal from scene content, and then freeze $\phi_\theta$ and use the resulting action stream $\alpha_{1:T-1}$ to train an auto-regressive video world model $G_\psi$.
Following prior latent-action work~\cite{edwards2019ilpo, bruce2024genie, schmidt2024lapo, lapa, gao2025adaworld}, we define the action as the change between two frames that is caused by the agent or the camera, and treat what can be inferred from a single frame, such as appearance, layout, and object identity, as scene content.
An overview of the training pipeline is provided in \cref{fig:method_overview}.
Once trained, $G_\psi$ supports two control modes: \emph{action transfer} and \emph{real-action control}.
\cref{sec:method_encoder} introduces the foundation-model-based action encoder, \cref{sec:method_contrastive} details its contrastive adaptation, and \cref{sec:method_world_model} describes the action-conditioned video world model.

\subsection{Frame-Pair Embeddings from a Foundation Model}
\label{sec:method_encoder}

We instantiate the action encoder $\phi_\theta$ with the encoder of V-JEPA~2~\cite{vjepa2}, a ViT-L foundation model pretrained on large-scale video.
Because it is pretrained to model how video content evolves across space and time, the encoder represents motion, not only appearance, which makes it a natural starting point from which to extract action representations.

\paragraph{Pair encoding.}
Given a frame pair $(o_t, o_{t+1})$, we stack the two frames into a two-frame clip and pass it through the encoder only; the V-JEPA~2 predictor is not used.
Since the encoder groups every two frames into one spatio-temporal tubelet, the clip yields a single temporal slice of $16 \times 16$ patch tokens, and each token jointly observes both frames from the first layer onward.
Motion is therefore fused into the representation from the start, rather than obtained by differencing two independently encoded frames.
We average-pool the output tokens, apply a fixed linear projection to $d{=}640$ dimensions, and $\ell_2$-normalize,
\[
    \alpha_t \;=\; \phi_\theta(o_t, o_{t+1}) \;=\; \frac{P\,\bar h_t}{\|P\,\bar h_t\|_2},
    \qquad
    \bar h_t \;=\; \frac{1}{N}\sum_{n=1}^{N} h_{t,n},
\]
where $h_{t,n}$ is the $n$-th output token and $P \in \mathbb{R}^{640 \times 1024}$ is a random projection that is fixed at initialization and never trained.
A random linear projection approximately preserves the geometry of the pooled embedding, so $P$ serves only to reduce dimensionality; all learning happens in the encoder.

The frozen embedding already encodes substantial action information (\cref{tab:action_probing}), but it is entangled with scene content: it leaks more scene information than latent-action models (\cref{tab:semantic_leakage}), and its action information degrades sharply on some domains.
A low-dimensional reconstruction bottleneck is the standard remedy for scene leakage, but it trades \emph{scene invariance against action expressiveness} and would discard the capacity that makes the pretrained space attractive in the first place.
We therefore adapt $\phi_\theta$ with a contrastive objective that reshapes its geometry without compressing it (\cref{sec:method_contrastive}).

\subsection{Contrastive Action Adaptation}
\label{sec:method_contrastive}

Building on the encoder $\phi_\theta$ of \cref{sec:method_encoder}, we now reshape its embedding space to better isolate action from scene while retaining the capacity of the pretrained representation.
We frame this adaptation around two properties that we want the action embedding space to satisfy: \emph{action expressiveness} and \emph{scene invariance}.
Without action labels, neither can be enforced directly, so we approximate both through a contrastive objective on positive and hard-negative pairs that are constructed from raw video.
To avoid overwriting the general-purpose features acquired during pretraining, we keep the pretrained weights $\theta$ frozen and attach low-rank adapters~\cite{hu2022lora} to the linear layers of every transformer block, yielding an update $\Delta\theta$ and the adapted encoder $\phi_{\theta+\Delta\theta}$.
With slight abuse of notation, we write $\phi_\theta$ for this adapted encoder in the remainder of the paper, with the understanding that $\theta$ subsumes $\Delta\theta$.

\paragraph{Pair construction.}
For each training sample we draw an anchor pair, a positive pair, and a hard-negative pair.
The \emph{anchor} is a frame pair $(o_t, o_{t+k})$ from a video, with a small random offset $k$.
The \emph{positive} $g(o_t, o_{t+k})$ should carry the same action in a different scene.
We construct it with \emph{flow-guided scene swapping}: we estimate the dense optical flow between the two frames with an off-the-shelf estimator~\cite{teed2020raft}, and threshold its magnitude to obtain the moving core region, i.e., the pixels that carry the action.
We cut out this region from both frames and paste it onto the same background frame taken from a different scene of the same dataset, translating it in both frames by a common random offset.
The pixels in motion are thus preserved exactly, while the scene, background, and absolute location of the motion are all replaced.
This construction is only valid when the camera is static. When the camera moves, as in egocentric, hand-held, or driving video, the background motion is itself part of the action and cannot be swapped; in this case the positive falls back to the anchor pair under a different photometric augmentation.
Both anchor and positive receive a color jitter that is shared between their two frames, so that inter-frame changes are not altered.
The \emph{hard negative} $(o_{t'}, o_{t'+k})$ is drawn from a non-overlapping time window of the same video: it shares the scene with the anchor but likely corresponds to a different action, so it cannot be told apart from the anchor by appearance.
We write $\alpha_t$, $\alpha_t^{+} = \phi_\theta\!\bigl(g(o_t, o_{t+k})\bigr)$, and $\alpha_t^{-} = \phi_\theta(o_{t'}, o_{t'+k})$ for the embeddings of the anchor, positive, and hard negative, respectively.

\paragraph{Contrastive objective.}
We organize these triplets into a candidate bank $\mathcal{Z}$ containing every anchor, positive, and hard negative in a batch of size $N$, and apply a symmetric InfoNCE objective~\cite{oord2018cpc} that pulls each anchor toward its matched positive and pushes it away from every other element of the bank in both directions,
\[
    \mathcal{L}_{\text{con}}
    \;=\;
    -\frac{1}{2N}\sum_{i=1}^{N}
    \left[
    \log
    \frac{\exp\!\bigl(\alpha_i^{\top}\alpha_i^{+}/\tau\bigr)}
         {\sum_{z \in \mathcal{Z}\setminus\{\alpha_i\}}\exp\!\bigl(\alpha_i^{\top} z/\tau\bigr)}
    +
    \log
    \frac{\exp\!\bigl(\alpha_i^{+\top}\alpha_i/\tau\bigr)}
         {\sum_{z \in \mathcal{Z}\setminus\{\alpha_i^{+}\}}\exp\!\bigl(\alpha_i^{+\top} z/\tau\bigr)}
    \right],
\]
where $\tau$ is a temperature.
The same-video hard negatives in $\mathcal{Z}$ make the objective non-trivial: in contrast, random in-batch negatives from unrelated scenes would be easily separated by appearance alone, allowing the encoder to satisfy the loss without disentangling action from scene.
The scene-swapped positives, in turn, prevent the encoder from separating the anchor from its hard negative by memorizing scene-specific cues, since the positive shares none of the anchor's background.

\paragraph{Inference.}
After adaptation, a latent action $\alpha_t$ is obtained by a single forward pass of $\phi_\theta$ on a frame pair. No labels, prompts, or trajectory-level optimization are required at inference.
The adapted encoder $\phi_\theta$ thus serves as a drop-in action encoder on arbitrary video, yielding the action stream $\alpha_{1:t}$ that conditions the video world model $G_\psi$ (\cref{sec:method_world_model}).

\begin{figure*}[t]
    \centering
    \includegraphics[width=\textwidth]{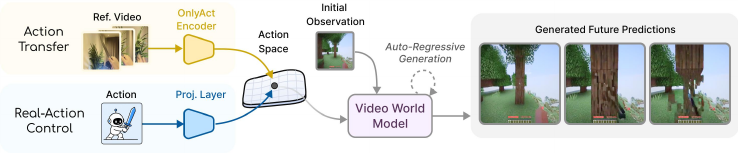}
    \caption{\textbf{Two control modes.} After pretraining the video world model on latent actions from the \mname{} encoder, the learned action space supports two modes. First, \emph{action transfer} extracts an action stream from a demonstration video and uses it to condition the world model from a new initial frame, reproducing the demonstrated action with no labels. Second, \emph{real-action control} learns a lightweight interface from real actions onto the learned latent-action space on a small action-labeled set, so that real actions drive the world model directly.}
    \label{fig:application}
\end{figure*}

\subsection{Action-Conditioned Video World Model}
\label{sec:method_world_model}

We obtain the action-conditioned video world model $G_\psi$ by adopting the autoregressive world model architecture of AdaWorld~\cite{gao2025adaworld} and replacing its latent actions by our action stream.
Given a trajectory $(o_1, \dots, o_T)$, we apply the frozen action encoder of \cref{sec:method_encoder} to consecutive frame pairs to form the \emph{action stream} $\alpha_{1:T-1}$ and use it as continuous action conditioning, with $G_\psi$ predicting each next frame $o_{t+1}$ from $(o_{1:t}, \alpha_{1:t})$.
Before entering $G_\psi$, our action embeddings are whitened by a fixed, parameter-free covariance equalization computed on the training data, while the baselines' latent actions enter unchanged; whitening a baseline in the same way leads to the same conclusions (Appendix~\ref{appendix:whitening_control}).
Otherwise, every action representation we compare uses exactly the same backbone, recipe, and data.

\paragraph{Action injection and training objective.}
Following AdaWorld, the action representation conditions the denoiser through two pathways: it is concatenated with the CLIP image embedding to form the cross-attention context, and concatenated with the context-length embedding into the vector condition, which a label-embedding MLP maps and adds to the timestep embedding.
One action vector is given per predicted transition, with no learned projection in between, and the action slots of the pretrained weights are zero-initialized.
$G_\psi$ predicts the next frame from up to six context frames, with the context length sampled uniformly from 1 to 6 during training, and is trained with the v-prediction objective under EDM noise scaling and loss weighting~\cite{karras2022edm}.
Each condition, including the action, is dropped with probability 0.1 to enable classifier-free guidance~\cite{ho2022cfg}.

\paragraph{Action transfer and real-action control.}
As illustrated in~\cref{fig:application}, the trained pipeline supports two control modes that share the same autoregressive rollout of the video world model $G_\psi$ but differ in how the conditioning action stream $\alpha_{1:T-1}$ is obtained.
For \emph{action transfer}, we extract $\alpha_{1:T-1}^{\mathrm{demo}}$ by running the frozen action encoder $\phi_\theta$ over a demonstration video and feed it to $G_\psi$ together with an initial frame $o_1$ from a different scene. The model then reproduces the demonstrated action in the new scene without any action labels.
For \emph{real-action control}, we fine-tune on a small action-labeled set.
Following~\cite{gao2026dreamdojo}, a small learned interface maps real actions into the latent-action space and replaces $\phi_\theta$ at the input of $G_\psi$: one vector per discrete action, initialized at the mean latent action of its labeled clips, or, for continuous actions, a two-layer MLP first fitted to regress the latent actions of the labeled clips.
The interface and the denoiser of $G_\psi$ are then fine-tuned end-to-end (Appendix~\ref{appendix:real_action_control}).
Because the latent-action space has already been shaped by large-scale unlabeled pretraining, only a modest amount of labeled data is required, and at inference time the mapped real actions serve as the conditioning stream.

\section{Experiments}
\label{sec:experiments}

We evaluate an action representation for action-conditioned video world models along three axes: (i) \textbf{action expressiveness} (\cref{exp:action_expressiveness}) --- how well the embedding distinguishes different actions; (ii) \textbf{scene invariance} (\cref{exp:scene_invariance}) --- whether the embedding avoids encoding scene or future-state content that a video generator could exploit as a shortcut; and (iii) \textbf{action-conditioned video generation quality} (\cref{exp:video_quality}) --- how well the embedding serves as a conditioning signal for video generation, measured by fidelity, transferability, and controllability.
On both seen and unseen benchmarks, \mname{} is substantially more expressive than prior annotation-free latent-action baselines, matches their scene invariance, and yields world models with better action transfer and controllability.
Implementation details are provided in Appendix~\ref{appendix:implementation_details}.

\subsection{Datasets and Baselines}

We evaluate \mname{} on datasets both seen and unseen during encoder adaptation. The action encoder is adapted on 168K unlabeled frame-pair samples from nine video sources spanning robot manipulation, human and egocentric video, camera navigation, and games (Appendix~\ref{appendix:contrastive_learning}). Among our evaluation datasets, only LIBERO~\cite{libero}, DROID~\cite{droid}, and SpatialVID~\cite{spatialvid} contribute videos to this set (marked $^{\circ}$ in the tables), and all other datasets are entirely unseen by the encoder.
For action probing and scene leakage, we use thirteen datasets with action annotations: nine robot-manipulation datasets (LIBERO, DROID, RLBench~\cite{james2019rlbench}, RoboCasa~\cite{robocasa}, DROID-IPRL~\cite{droid}, MetaWorld~\cite{metaworld}, MUTEX~\cite{mutex}, Robosuite~\cite{robosuite}, and RoboDesk~\cite{robodesk}), three camera-navigation datasets (SpatialVID, ARKitScenes~\cite{arkitscenes}, and RealEstate10K~\cite{realestate}), and Procgen~\cite{cobbe2019procgen} with 15 discrete actions. For action-conditioned video generation, we evaluate on LIBERO, RLBench, SpatialVID, and nuScenes~\cite{nuscenes}, and further fine-tune with real actions on DMLab~\cite{dmlab} and Minecraft from the TECO datasets~\cite{teco}, Habitat~\cite{habitat} in HM3D scenes~\cite{hm3d}, and nuScenes.

We compare against two families of baselines. The first is latent-action models, namely DreamDojo~\cite{gao2026dreamdojo} and AdaWorld~\cite{gao2025adaworld}. Since their publicly released weights are not trained on our data, we retrain both on the same 168K training pairs as our encoder for a fair comparison (marked $^\dagger$).
The second is the frozen V-JEPA~2 encoder~\cite{vjepa2}, embedded in the same way as \mname{} (\cref{sec:method_encoder}) but without contrastive adaptation, which isolates the effect of our adaptation from that of the pretrained backbone.

\begin{table*}[t]
\centering
\caption{\textbf{Linear probing of action information.} A linear layer predicts the ground-truth action from each frozen representation. We report test $R^2$ on held-out scenes (LIBERO, RLBench: held-out demonstrations of training scenes; Procgen: macro-F1 over 15 discrete actions), averaged over 3 probe seeds (std $\le$ 0.03). MetaWorld actions are clipped to $[-1, 1]$. $^{\circ}$Videos seen by our encoder during training, without action labels; all other datasets are entirely unseen. $^\dagger$Retrained on the same 168K training pairs as our encoder. V-JEPA~2 (frozen): our backbone, embedded identically but without contrastive adaptation. Avg.: mean $R^2$ over the 12 continuous-action datasets. $\Delta$: ours minus the best other method. \best{Bold}: best; \second{underline}: second best.}
\label{tab:action_probing}
\setlength{\tabcolsep}{4.5pt}
\renewcommand{\arraystretch}{1.12}
\resizebox{\textwidth}{!}{%
\begin{tabular}{@{}lccccccccc@{\hspace{9pt}}ccc@{\hspace{9pt}}c@{\hspace{9pt}}c@{}}
\toprule
 & \multicolumn{9}{c}{\textit{Robot manipulation}} & \multicolumn{3}{c}{\textit{Camera navigation}} & \multicolumn{1}{c}{\textit{Game (F1)}} & \\
\cmidrule(lr){2-10} \cmidrule(lr){11-13} \cmidrule(lr){14-14}
\textbf{Representation} & LIBERO$^{\circ}$ & DROID$^{\circ}$ & RLBench & RoboCasa & DROID-IPRL & MetaWorld & MUTEX & Robosuite & RoboDesk & SpatialVID$^{\circ}$ & ARKitScenes & RE10K & Procgen & Avg. \\
\midrule
\textcolor{gray}{Random} & \textcolor{gray}{$-$0.032} & \textcolor{gray}{$-$0.112} & \textcolor{gray}{$-$0.021} & \textcolor{gray}{$-$0.100} & \textcolor{gray}{$-$0.094} & \textcolor{gray}{$-$0.062} & \textcolor{gray}{$-$0.156} & \textcolor{gray}{$-$0.093} & \textcolor{gray}{$-$0.072} & \textcolor{gray}{$-$0.092} & \textcolor{gray}{$-$0.090} & \textcolor{gray}{$-$0.074} & \textcolor{gray}{0.073} & \textcolor{gray}{$-$0.083} \\
\midrule
AdaWorld$^\dagger$~\cite{gao2025adaworld} & 0.199 & 0.011 & 0.136 & $-$0.154 & 0.009 & $-$0.045 & 0.191 & 0.121 & 0.170 & 0.108 & 0.006 & 0.046 & 0.063 & 0.067 \\
DreamDojo$^\dagger$~\cite{gao2026dreamdojo} & 0.324 & \second{0.071} & 0.246 & \second{0.034} & 0.162 & \second{0.156} & 0.425 & 0.319 & 0.268 & 0.266 & 0.087 & 0.173 & 0.092 & 0.211 \\
\midrule
V-JEPA~2 (frozen)~\cite{vjepa2} & \best{0.718} & $-$0.131 & \second{0.556} & $-$0.368 & \second{0.186} & 0.144 & \second{0.527} & \second{0.448} & \second{0.520} & \second{0.301} & \best{0.665} & \second{0.598} & \second{0.145} & \second{0.347} \\
\rowcolor{oursblue}
\textbf{\mname} (ours) & \second{0.634} & \best{0.203} & \best{0.566} & \best{0.322} & \best{0.279} & \best{0.277} & \best{0.566} & \best{0.695} & \best{0.581} & \best{0.560} & \second{0.588} & \best{0.609} & \best{0.200} & \best{0.490} \\
\multicolumn{1}{@{}r}{\footnotesize $\Delta$} & \footnotesize \textcolor{deltaneg}{$-$0.08} & \footnotesize \textcolor{deltapos}{+0.13} & \footnotesize \textcolor{deltapos}{+0.01} & \footnotesize \textcolor{deltapos}{+0.29} & \footnotesize \textcolor{deltapos}{+0.09} & \footnotesize \textcolor{deltapos}{+0.12} & \footnotesize \textcolor{deltapos}{+0.04} & \footnotesize \textcolor{deltapos}{+0.25} & \footnotesize \textcolor{deltapos}{+0.06} & \footnotesize \textcolor{deltapos}{+0.26} & \footnotesize \textcolor{deltaneg}{$-$0.08} & \footnotesize \textcolor{deltapos}{+0.01} & \footnotesize \textcolor{deltapos}{+0.05} & \footnotesize \textcolor{deltapos}{+0.14} \\
\bottomrule
\end{tabular}}
\end{table*}

\subsection{Action Expressiveness}
\label{exp:action_expressiveness}

To evaluate whether the representation captures rich and discriminative action dynamics, we conduct action probing experiments. Specifically, we train a linear probe to predict ground-truth actions from each representation, and report the results in \cref{tab:action_probing} (details in Appendix~\ref{appendix:action_expressiveness}).

As shown in \cref{tab:action_probing}, \mname{} captures more expressive and discriminative action dynamics than the baselines on 11 of the 13 benchmarks, raising the average $R^2$ from 0.211 for the strongest latent-action model to 0.490.
The frozen V-JEPA~2 encoder already outperforms the latent-action models on average, confirming that the pretrained foundation model provides a strong starting point for action representation. However, its action information is unreliable across domains, and on DROID and RoboCasa its $R^2$ even falls below that of the latent-action models.
Contrastive adaptation removes these failure cases and improves the average $R^2$ from 0.347 to 0.490, with large gains also on datasets unseen during adaptation (e.g., +0.69 on RoboCasa and +0.25 on Robosuite). We attribute this improved transferability to the broad spatio-temporal priors of the pretrained foundation model, which the adaptation preserves rather than overwrites.

High action prediction accuracy alone, however, does not guarantee a representation suitable for action-conditioned video generation, as we analyze next.

\begin{table*}[t]
\centering
\caption{\textbf{Scene leakage.} A decoder with the same architecture for every representation (trained with L1 + 0.5$\,\times\,$LPIPS, 10 epochs, $128{\times}128$) reconstructs the post-action frame from the representation and is evaluated on scenes held out from its training; values are averaged over 4 decoder seeds. Per-dataset columns: held-out LPIPS$\uparrow$ (higher = less scene information recoverable). Last three columns: LPIPS$\uparrow$, SSIM$\downarrow$ and L1$\uparrow$ averaged over the 13 datasets. A representation that carries no information at all also leaks nothing, so read this table together with Table~\ref{tab:action_probing}. Markers and baselines as in Table~\ref{tab:action_probing}. $\Delta$: ours minus the best other method (green: ours leaks less). \best{Bold}: least leakage; \second{underline}: second least.}
\label{tab:semantic_leakage}
\setlength{\tabcolsep}{4.5pt}
\renewcommand{\arraystretch}{1.12}
\resizebox{\textwidth}{!}{%
\begin{tabular}{@{}lccccccccc@{\hspace{9pt}}ccc@{\hspace{9pt}}c@{\hspace{9pt}}ccc@{}}
\toprule
 & \multicolumn{9}{c}{\textit{Robot manipulation}} & \multicolumn{3}{c}{\textit{Camera navigation}} & \multicolumn{1}{c}{\textit{Game}} & \multicolumn{3}{c}{\textit{Average}} \\
\cmidrule(lr){2-10} \cmidrule(lr){11-13} \cmidrule(lr){14-14} \cmidrule(lr){15-17}
\textbf{Representation} & LIBERO$^{\circ}$ & DROID$^{\circ}$ & RLBench & RoboCasa & DROID-IPRL & MetaWorld & MUTEX & Robosuite & RoboDesk & SpatialVID$^{\circ}$ & ARKitScenes & RE10K & Procgen & LPIPS$\uparrow$ & SSIM$\downarrow$ & L1$\uparrow$ \\
\midrule
AdaWorld$^\dagger$~\cite{gao2025adaworld} & 0.288 & 0.314 & \second{0.273} & 0.363 & \second{0.320} & \second{0.136} & \second{0.211} & 0.307 & \best{0.127} & 0.366 & \second{0.405} & 0.369 & 0.325 & 0.293 & \second{0.374} & \second{0.256} \\
DreamDojo$^\dagger$~\cite{gao2026dreamdojo} & \second{0.306} & \second{0.315} & \best{0.283} & 0.375 & 0.317 & \best{0.137} & \best{0.212} & \second{0.308} & \second{0.119} & 0.379 & 0.405 & 0.385 & \second{0.347} & \second{0.299} & \best{0.369} & \best{0.260} \\
\midrule
V-JEPA~2 (frozen)~\cite{vjepa2} & 0.257 & 0.289 & 0.216 & \best{0.480} & 0.247 & 0.066 & 0.130 & 0.282 & 0.069 & \second{0.402} & 0.384 & \best{0.442} & 0.311 & 0.275 & 0.478 & 0.200 \\
\rowcolor{oursblue}
\textbf{\mname} (ours) & \best{0.314} & \best{0.328} & 0.269 & \second{0.411} & \best{0.322} & 0.105 & 0.177 & \best{0.317} & 0.100 & \best{0.407} & \best{0.411} & \second{0.436} & \best{0.352} & \best{0.304} & 0.391 & 0.255 \\
\multicolumn{1}{@{}r}{\footnotesize $\Delta$} & \footnotesize \textcolor{deltapos}{+0.008} & \footnotesize \textcolor{deltapos}{+0.014} & \footnotesize \textcolor{deltaneg}{$-$0.014} & \footnotesize \textcolor{deltaneg}{$-$0.069} & \footnotesize \textcolor{deltapos}{+0.002} & \footnotesize \textcolor{deltaneg}{$-$0.032} & \footnotesize \textcolor{deltaneg}{$-$0.035} & \footnotesize \textcolor{deltapos}{+0.009} & \footnotesize \textcolor{deltaneg}{$-$0.027} & \footnotesize \textcolor{deltapos}{+0.005} & \footnotesize \textcolor{deltapos}{+0.005} & \footnotesize \textcolor{deltaneg}{$-$0.006} & \footnotesize \textcolor{deltapos}{+0.005} & \footnotesize \textcolor{deltapos}{+0.005} & \footnotesize \textcolor{deltaneg}{+0.021} & \footnotesize \textcolor{deltaneg}{$-$0.004} \\
\bottomrule
\end{tabular}}
\end{table*}

\subsection{Scene Invariance}
\label{exp:scene_invariance}

Since action embeddings are inferred from consecutive frames, they may inadvertently encode future-state information that a video model can exploit as a shortcut, directly reconstructing future content rather than learning genuine action-conditioned dynamics and thereby hurting both controllability and transferability.

\begin{figure*}[t]
    \centering
    \resizebox{\textwidth}{!}{%
        \includegraphics[height=4cm, trim=00 10 20 10, clip]{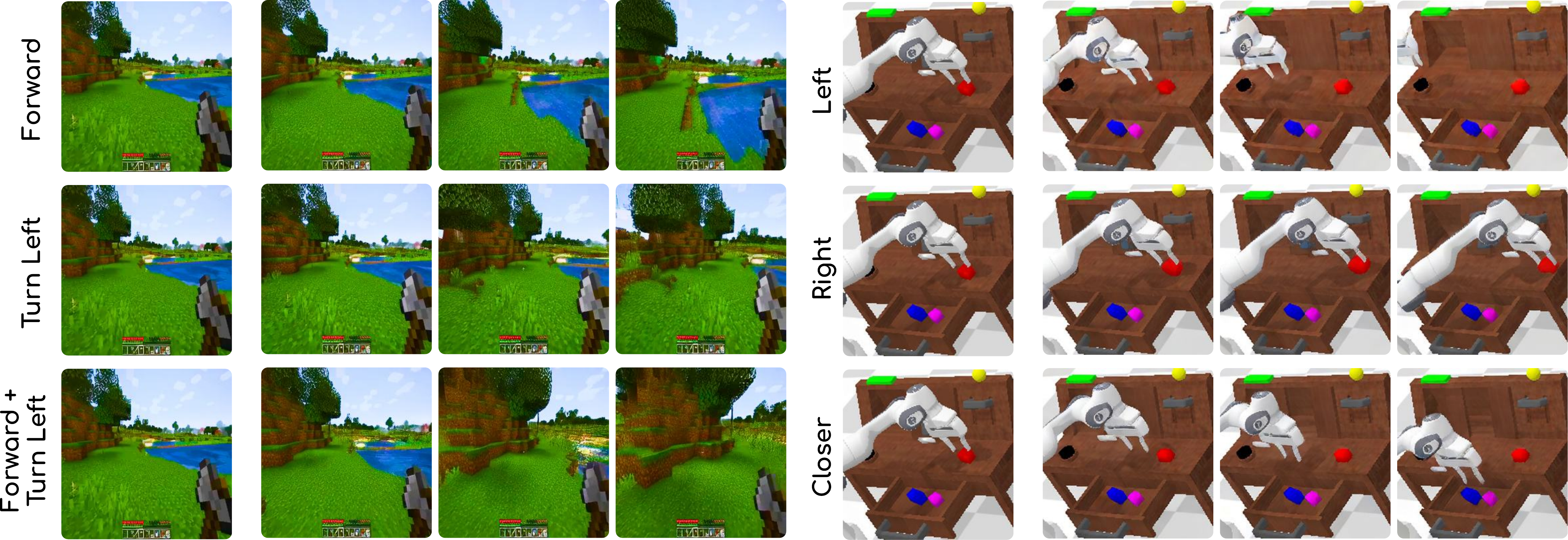}%
        \hspace{0.4cm}%
        \raisebox{-0.25cm}{\includegraphics[height=4.5cm, trim=0 0 0 0, clip]{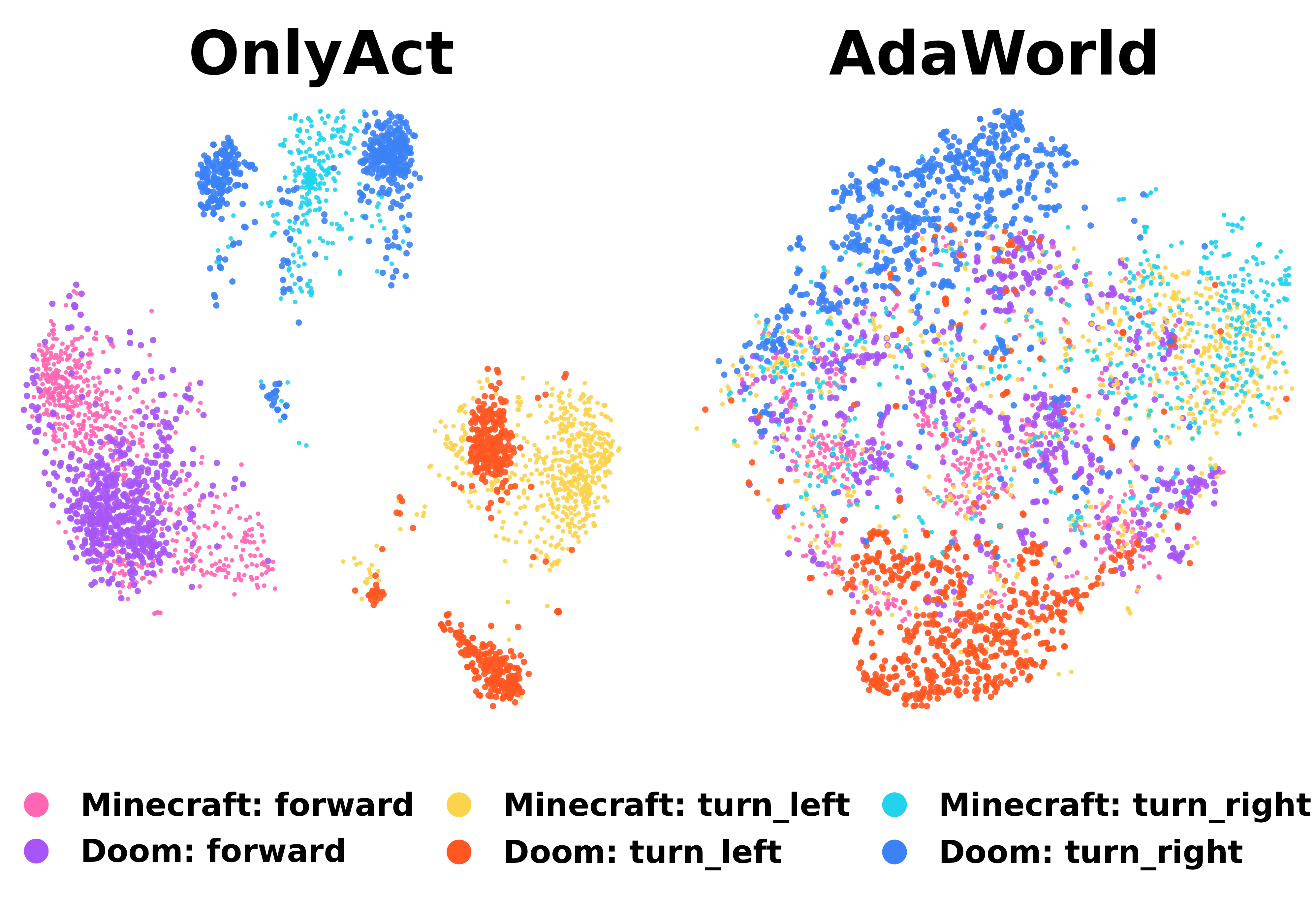}}%
    }
    \vspace{-1mm}
    \caption{\textbf{Real-action control}. \textbf{Left:} \mname{} learns to generate videos that faithfully respond to different actions, \textbf{Right:} comparison of action representations between \mname{} and AdaWorld using t-SNE.}
    \label{fig:different-action-tsne}
\end{figure*}

To quantify how much scene information is encoded in the action embeddings, we train a decoder to reconstruct the future frame from the action embedding alone and evaluate it on held-out scenes. A more scene-invariant embedding should yield worse reconstruction. The results in \cref{tab:semantic_leakage} show that the frozen V-JEPA~2 embedding leaks the most scene information (average LPIPS 0.275). With contrastive adaptation, however, \mname{} matches the scene invariance of latent-action models, attaining the highest average LPIPS (0.304) and comparable SSIM and L1, while retaining a substantially more expressive action representation. The adaptation therefore improves expressiveness and scene invariance together, without relying on a bottleneck. The right side of \cref{fig:different-action-tsne} further shows that \mname{} embeddings form meaningful clusters for the same actions across different datasets. The decoder structure is described in Appendix~\ref{appendix:scene_invariance}.

\subsection{Action-Conditioned Video Generation Quality}
\label{exp:video_quality}

The ultimate goal of the learned action representation is to provide a general conditioning signal for action-conditioned video generation. We hypothesize that embeddings that are both expressive and scene-invariant enable a video model to learn action-conditioned dynamics more effectively, generalize across domains, and transfer to downstream tasks after lightweight fine-tuning with real actions.

To evaluate this hypothesis, we train the same action-conditioned autoregressive world model (\cref{sec:method_world_model}) with each action representation as input, using the same backbone, recipe, and data for all of them. We assess generation quality along three sub-axes: \emph{fidelity}, \emph{transferability}, and \emph{controllability}. Fidelity is measured by the PSNR of generations driven by the actions inferred from the clip itself (Self PSNR). Transferability is measured by driving a different scene with the actions of a source clip and computing the cosine between the optical-flow direction~\cite{farneback2003} of the generated video and that of the source (Transfer Dir), where 0 means unrelated and 1 means the same direction. Since a representation that leaks scene content can reach a high Self PSNR by reconstructing the future frame but cannot carry its action to a new scene, Transfer Dir is the more direct test of an action representation.

\begin{figure*}[t]
    \centering
    \setlength{\tabcolsep}{1.5pt}
    \renewcommand{\arraystretch}{1}%
    \newlength{\qatimgw}\setlength{\qatimgw}{0.321\textwidth}
    \newlength{\qatimgh}\setlength{\qatimgh}{0.5\qatimgw}
    \begin{tabular}{@{}c@{\hspace{1pt}}c@{\hspace{3pt}}c@{\hspace{3pt}}c@{}}
        &
        {\comfortaa\fontsize{5}{6}\selectfont Same-source transfer (CALVIN)} &
        {\comfortaa\fontsize{5}{6}\selectfont Cross-source transfer (DOOM $\to$ Minecraft)} &
        {\comfortaa\fontsize{5}{6}\selectfont In-the-wild demonstration} \\[0pt]
        \vbox to \qatimgh{%
            \hsize=0.7em
            \vfil
            \hbox to \hsize{\hfil\rotatebox{90}{{\comfortaa\fontsize{5}{6}\selectfont Demo.}}\hfil}%
            \vfil\vfil
            \hbox to \hsize{\hfil\rotatebox{90}{{\comfortaa\fontsize{5}{6}\selectfont Gen.}}\hfil}%
            \vfil
        }%
        & \includegraphics[width=\qatimgw]{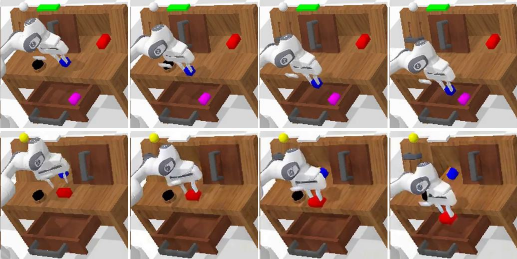}
        & \includegraphics[width=\qatimgw]{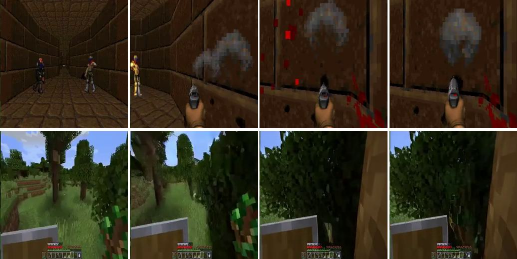}
        & \includegraphics[width=\qatimgw]{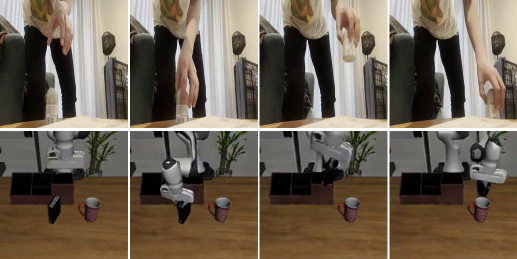} \\
    \end{tabular}
    \caption{\textbf{Action transfer.} Latent actions extracted from the demonstration frames (top) condition the video world model generations (bottom). \mname{} faithfully reproduces the demonstrated action in the new target scene, e.g. in the cross-source example it turns right and then performs the attack action. Best seen as a video in the supplementary material.}
    \label{fig:qual-action-transfer}
\end{figure*}

\begin{table*}[t]
\centering
\caption{\textbf{Action-conditioned world model.} The same world model (AdaWorld backbone, same recipe and data) is trained with each action representation. \emph{Self}: PSNR$\uparrow$ (dB) when driven by the clip's own actions. \emph{Transfer}: a source clip's actions drive a different scene; Dir$\uparrow$ is the cosine between the optical-flow direction of the generated video and of the source (0: unrelated; 1: same). $^{\circ}$Seen by our encoder without action labels. $^\dagger$Retrained on our data. $\Delta$: ours minus the best baseline. \best{Bold}: best; \second{underline}: second best (no-action row excluded).}
\label{tab:main_video_quality}
\setlength{\tabcolsep}{5pt}
\renewcommand{\arraystretch}{1.12}
\resizebox{\textwidth}{!}{%
\begin{tabular}{@{}lcccc@{\hspace{10pt}}cc@{\hspace{10pt}}cccc@{}}
\toprule
 & \multicolumn{4}{c}{\textit{Robot manipulation}} & \multicolumn{2}{c}{\textit{Camera navigation}} & \multicolumn{4}{c}{\textit{Driving}} \\
\cmidrule(lr){2-5} \cmidrule(lr){6-7} \cmidrule(lr){8-11}
 & \multicolumn{2}{c}{LIBERO$^{\circ}$} & \multicolumn{2}{c}{RLBench} & \multicolumn{2}{c}{SpatialVID$^{\circ}$} & \multicolumn{2}{c}{nuScenes dev.} & \multicolumn{2}{c}{nuScenes conf.} \\
\cmidrule(lr){2-3} \cmidrule(lr){4-5} \cmidrule(lr){6-7} \cmidrule(lr){8-9} \cmidrule(lr){10-11}
\textbf{Action representation} & {\small Self PSNR$\uparrow$} & {\small Transfer Dir$\uparrow$} & {\small Self PSNR$\uparrow$} & {\small Transfer Dir$\uparrow$} & {\small Self PSNR$\uparrow$} & {\small Transfer Dir$\uparrow$} & {\small Self PSNR$\uparrow$} & {\small Transfer Dir$\uparrow$} & {\small Self PSNR$\uparrow$} & {\small Transfer Dir$\uparrow$} \\
\midrule
\textcolor{gray}{No action} & \textcolor{gray}{18.41} & \textcolor{gray}{0.186} & \textcolor{gray}{18.51} & \textcolor{gray}{$-$0.054} & \textcolor{gray}{15.42} & \textcolor{gray}{$-$0.188} & \textcolor{gray}{18.60} & \textcolor{gray}{$-$0.092} & \textcolor{gray}{19.10} & \textcolor{gray}{0.398} \\
\midrule
AdaWorld$^\dagger$~\cite{gao2025adaworld} & 18.90 & 0.162 & \second{18.32} & $-$0.109 & 15.52 & $-$0.274 & 18.44 & $-$0.142 & \second{19.46} & 0.432 \\
DreamDojo$^\dagger$~\cite{gao2026dreamdojo} & \second{19.95} & \second{0.803} & 18.08 & \second{0.572} & \best{16.40} & \second{0.480} & \best{18.57} & \second{0.574} & \best{19.67} & \second{0.571} \\
\midrule
\rowcolor{oursblue}
\textbf{\mname} (ours) & \best{20.32} & \best{0.917} & \best{18.63} & \best{0.686} & \second{16.10} & \best{0.608} & \second{18.56} & \best{0.659} & 19.38 & \best{0.603} \\
\multicolumn{1}{@{}r}{\footnotesize $\Delta$} & \footnotesize \textcolor{deltapos}{+0.37} & \footnotesize \textcolor{deltapos}{+0.114} & \footnotesize \textcolor{deltapos}{+0.31} & \footnotesize \textcolor{deltapos}{+0.113} & \footnotesize \textcolor{deltaneg}{$-$0.31} & \footnotesize \textcolor{deltapos}{+0.128} & \footnotesize \textcolor{deltaneg}{$-$0.01} & \footnotesize \textcolor{deltapos}{+0.085} & \footnotesize \textcolor{deltaneg}{$-$0.29} & \footnotesize \textcolor{deltapos}{+0.032} \\
\bottomrule
\end{tabular}}
\end{table*}

\begin{table*}[t]
\centering
\caption{\textbf{Downstream world-model fine-tuning: controllability and fidelity.} Each pretrained world model from Table~\ref{tab:main_video_quality} is fine-tuned on a small labeled set of a target domain (all unseen by our encoder, which was trained on VPT Minecraft recordings, not these TECO videos). \textbf{(a) Controllability} after 800 updates. Follow$\uparrow$: accuracy of an inverse-dynamics classifier at recognizing the commanded action in the generated frame (chance 0.333; \emph{Real frames}: the same classifier on real frames); on nuScenes, Spearman correlation between commanded and generated forward motion (chance 0). Control$\uparrow$: PSNR under the commanded action minus under other actions (dB; 0 = action ignored). \textbf{(b) Fidelity}: PSNR$\uparrow$ (dB) after 0, 200 and 800 updates (0: interface initialized from the labeled clips, no gradient update). Markers as in Table~\ref{tab:main_video_quality}. $\Delta$: ours minus the best baseline; grey when the 95\% paired bootstrap interval contains 0. \best{Bold}: best; \second{underline}: second best.}
\label{tab:downstream_finetune}
\setlength{\tabcolsep}{4pt}
\renewcommand{\arraystretch}{1.12}
\resizebox{\textwidth}{!}{%
\begin{tabular}{@{}lccc@{\hspace{8pt}}ccc@{\hspace{8pt}}ccc@{\hspace{8pt}}ccc@{}}
\toprule
 & \multicolumn{3}{c}{DMLab} & \multicolumn{3}{c}{Minecraft} & \multicolumn{3}{c}{Habitat} & \multicolumn{3}{c}{nuScenes} \\
\cmidrule(lr){2-4} \cmidrule(lr){5-7} \cmidrule(lr){8-10} \cmidrule(lr){11-13}
\multicolumn{13}{@{}l}{\textbf{(a) Controllability} after 800 updates} \\
\textbf{Action representation} & {\small Follow$\uparrow$} & \multicolumn{2}{c}{{\small Control$\uparrow$}} & {\small Follow$\uparrow$} & \multicolumn{2}{c}{{\small Control$\uparrow$}} & {\small Follow$\uparrow$} & \multicolumn{2}{c}{{\small Control$\uparrow$}} & {\small Follow$\uparrow$} & \multicolumn{2}{c}{{\small Control$\uparrow$}} \\
\midrule
\textcolor{gray}{Real frames} & \textcolor{gray}{0.997} & \multicolumn{2}{c}{\textcolor{gray}{--}} & \textcolor{gray}{0.900} & \multicolumn{2}{c}{\textcolor{gray}{--}} & \textcolor{gray}{0.787} & \multicolumn{2}{c}{\textcolor{gray}{--}} & \textcolor{gray}{--} & \multicolumn{2}{c}{\textcolor{gray}{--}} \\
AdaWorld$^\dagger$~\cite{gao2025adaworld} & 0.508 & \multicolumn{2}{c}{4.04} & 0.647 & \multicolumn{2}{c}{6.88} & 0.527 & \multicolumn{2}{c}{3.04} & 0.628 & \multicolumn{2}{c}{0.04} \\
DreamDojo$^\dagger$~\cite{gao2026dreamdojo} & \second{0.863} & \multicolumn{2}{c}{\second{7.99}} & \second{0.787} & \multicolumn{2}{c}{\second{7.59}} & \second{0.749} & \multicolumn{2}{c}{\second{7.33}} & \second{0.722} & \multicolumn{2}{c}{\second{0.15}} \\
\midrule
\rowcolor{oursblue}
\textbf{\mname} (ours) & \best{0.901} & \multicolumn{2}{c}{\best{8.95}} & \best{0.821} & \multicolumn{2}{c}{\best{7.81}} & \best{0.767} & \multicolumn{2}{c}{\best{7.35}} & \best{0.928} & \multicolumn{2}{c}{\best{0.64}} \\
\multicolumn{1}{@{}r}{\footnotesize $\Delta$} & \footnotesize \textcolor{deltapos}{+0.038} & \multicolumn{2}{c}{\footnotesize \textcolor{deltapos}{+0.96}} & \footnotesize \textcolor{deltapos}{+0.034} & \multicolumn{2}{c}{\footnotesize \textcolor{deltapos}{+0.22}} & \footnotesize \textcolor{deltapos}{+0.018} & \multicolumn{2}{c}{\footnotesize \textcolor{gray}{+0.02}} & \footnotesize \textcolor{deltapos}{+0.207} & \multicolumn{2}{c}{\footnotesize \textcolor{deltapos}{+0.49}} \\
\midrule\midrule
\multicolumn{13}{@{}l}{\textbf{(b) Fidelity}, PSNR$\uparrow$ (dB) after 0 / 200 / 800 updates} \\
\textbf{Action representation} & {\small 0} & {\small 200} & {\small 800} & {\small 0} & {\small 200} & {\small 800} & {\small 0} & {\small 200} & {\small 800} & {\small 0} & {\small 200} & {\small 800} \\
\midrule
AdaWorld$^\dagger$~\cite{gao2025adaworld} & 18.77 & 21.35 & 22.34 & \best{23.07} & 23.36 & 23.46 & 18.44 & 18.95 & 20.99 & \best{24.59} & \best{24.85} & \best{24.77} \\
DreamDojo$^\dagger$~\cite{gao2026dreamdojo} & \second{19.14} & \second{22.56} & \second{23.08} & \second{22.80} & \second{23.48} & \second{23.49} & \second{19.24} & \second{23.00} & \best{23.78} & \second{24.53} & \second{24.83} & \best{24.77} \\
\midrule
\rowcolor{oursblue}
\textbf{\mname} (ours) & \best{19.28} & \best{22.82} & \best{23.11} & 22.57 & \best{23.59} & \best{23.54} & \best{20.56} & \best{23.27} & \second{23.74} & \best{24.59} & \second{24.83} & \second{24.76} \\
\multicolumn{1}{@{}r}{\footnotesize $\Delta$} & \footnotesize \textcolor{deltapos}{+0.14} & \footnotesize \textcolor{deltapos}{+0.26} & \footnotesize \textcolor{gray}{+0.03} & \footnotesize \textcolor{deltaneg}{$-$0.50} & \footnotesize \textcolor{deltapos}{+0.11} & \footnotesize \textcolor{gray}{+0.05} & \footnotesize \textcolor{deltapos}{+1.31} & \footnotesize \textcolor{deltapos}{+0.27} & \footnotesize \textcolor{gray}{$-$0.04} & \footnotesize \textcolor{gray}{0.00} & \footnotesize \textcolor{gray}{$-$0.02} & \footnotesize \textcolor{gray}{0.00} \\
\bottomrule
\end{tabular}}
\end{table*}
The results in \cref{tab:main_video_quality} show that \mname{} achieves the best Transfer Dir on all five evaluation settings, improving over DreamDojo by $+0.03$ to $+0.13$, including on RLBench and nuScenes, which are unseen by our encoder, while AdaWorld's latent actions stay close to or below the no-action reference. Fidelity is comparable across methods: \mname{} has the highest Self PSNR on LIBERO and RLBench and is within 0.31\,dB of the best method on SpatialVID and nuScenes.
In addition, \cref{fig:qual-action-transfer} and further video results in the supplementary material show that the action-transfer control mode (\cref{sec:method_world_model}) faithfully reproduces a demonstrated action in a new target scene, including under cross-source transfer and from in-the-wild demonstrations.

Controllability is evaluated by fine-tuning each pretrained world model on a small action-labeled set of a target domain (real-action control, \cref{sec:method_world_model}; details in Appendix~\ref{appendix:real_action_control}). We report Follow, the accuracy of an inverse-dynamics classifier at recognizing the commanded action in the generated frame (on nuScenes, the Spearman correlation between commanded and generated forward motion), and Control, the PSNR gap between generations under the commanded action and under other actions.
As shown in \cref{tab:downstream_finetune}, \mname{} yields the most controllable world model on all four domains, with the largest gain on nuScenes (Follow 0.928 vs.\ 0.722); on Habitat, the Control gain is within noise. \mname{} also reaches the highest fidelity early in fine-tuning on DMLab and Habitat, while after 800 updates fidelity is on par across representations. This indicates that the action space shaped by unlabeled adaptation transfers effectively to real-action fine-tuning, as further illustrated qualitatively on the left of \cref{fig:different-action-tsne}.

\section{Conclusion}

We presented \mname{}, a scalable approach to action representation learning for video world models.
Previous latent-action models typically learn action spaces from scratch using unlabeled videos, often relying on restrictive bottlenecks to disentangle actions from scene content. In contrast, \mname{} adapts a pretrained foundation model into an action encoder and uses a label-free contrastive objective, with flow-guided scene-swapped positives and same-video hard negatives, to steer its embeddings towards action-centric representations, without imposing such a bottleneck.
We show that the resulting action representations are substantially more expressive than those learned by prior annotation-free latent-action methods while matching their scene invariance, and that they enable world models with better action transfer and controllability on new domains.

\begin{ack}
This work was supported by the Clarendon Scholarship at the University of Oxford. The authors declare no competing interests.
\end{ack}

{
\small
\bibliographystyle{plain}
\bibliography{ref/main}%
}

\newpage
\newpage

\appendix
\section*{Appendix}
\addcontentsline{toc}{part}{Appendix}   %

Qualitative samples are shown in the local HTML in the supplementary materials.

\begingroup
  \etocsetnexttocdepth{subsection}     %
  \etocsettocstyle{}{}
  \localtableofcontents
\endgroup

\section{Additional Experiments}

\subsection{Action-Transfer Videos}
\label{appendix:action_transfer_videos}
Extensive qualitative videos for the action-transfer control mode (\cref{sec:method_world_model}) are provided in the supplementary material, covering same-source action transfer, cross-source action transfer, and in-the-wild demonstrations. \cref{fig:appendix-action-transfer} shows a few such examples; please see the supplementary HTML for the full set.

\begin{figure*}[]
    \centering
    \setlength{\tabcolsep}{1.5pt}
    \renewcommand{\arraystretch}{1}%
    \newlength{\qatappimgw}\setlength{\qatappimgw}{0.321\textwidth}
    \newlength{\qatappimgh}\setlength{\qatappimgh}{0.5\qatappimgw}
    \newcommand{\qatapplabel}{%
        \vbox to \qatappimgh{%
            \hsize=0.7em
            \vfil
            \hbox to \hsize{\hfil\rotatebox{90}{{\comfortaa\fontsize{5}{6}\selectfont Demo.}}\hfil}%
            \vfil\vfil
            \hbox to \hsize{\hfil\rotatebox{90}{{\comfortaa\fontsize{5}{6}\selectfont Gen.}}\hfil}%
            \vfil
        }%
    }
    \newcommand{\qatapphdr}[1]{\multicolumn{4}{@{}c@{}}{{\comfortaa\fontsize{6}{7}\selectfont #1}}}
    \begin{tabular}{@{}c@{\hspace{1pt}}c@{\hspace{3pt}}c@{\hspace{3pt}}c@{}}
        \qatapphdr{Same-source action transfer: LIBERO (in-domain demonstration \& target)} \\
        \qatapplabel
        & \includegraphics[width=\qatappimgw]{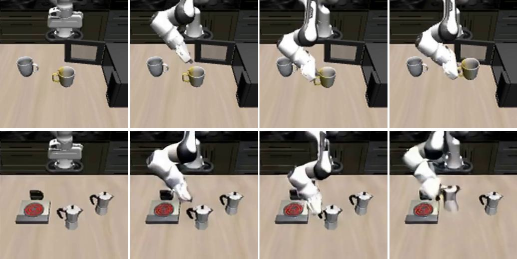}
        & \includegraphics[width=\qatappimgw]{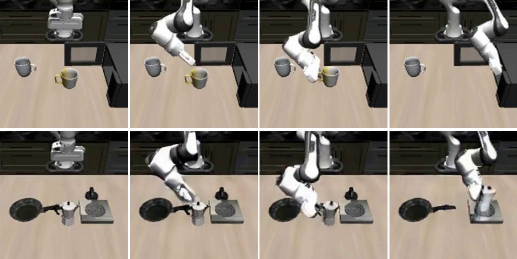}
        & \includegraphics[width=\qatappimgw]{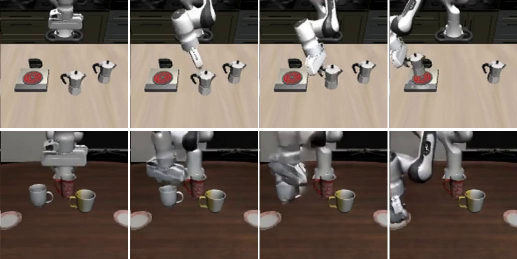} \\[4pt]
        \qatapphdr{Same-source action transfer: CALVIN (in-domain demonstration \& target)} \\
        \qatapplabel
        & \includegraphics[width=\qatappimgw]{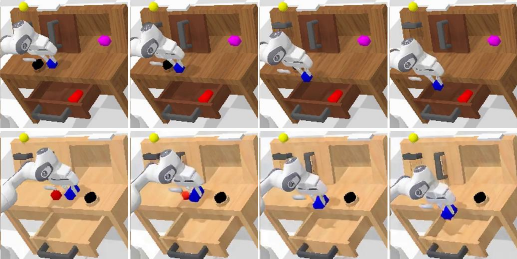}
        & \includegraphics[width=\qatappimgw]{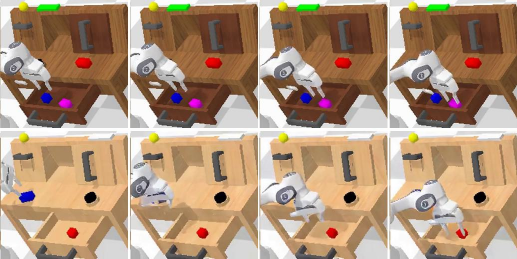}
        & \includegraphics[width=\qatappimgw]{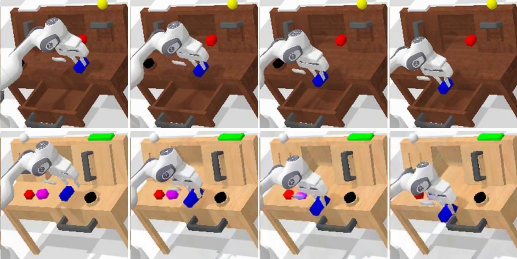} \\[4pt]
        \qatapphdr{Same-source action transfer: Minecraft (in-domain demonstration \& target)} \\
        \qatapplabel
        & \includegraphics[width=\qatappimgw]{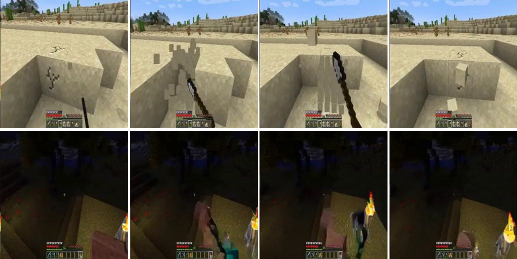}
        & \includegraphics[width=\qatappimgw]{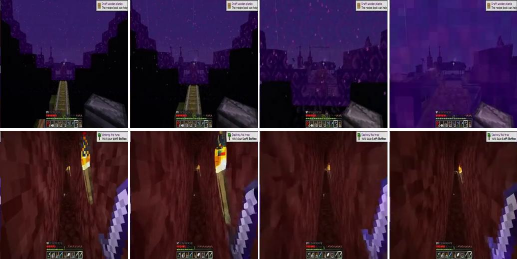}
        & \includegraphics[width=\qatappimgw]{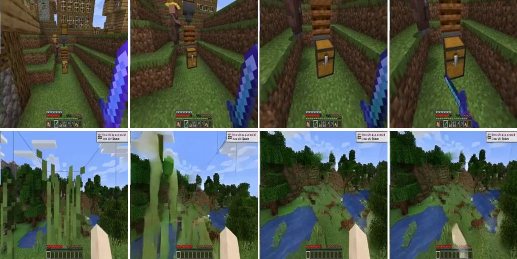} \\[4pt]
        \qatapphdr{In-the-wild demonstration (real-world captured demonstration $\to$ in-domain target)} \\
        \qatapplabel
        & \includegraphics[width=\qatappimgw]{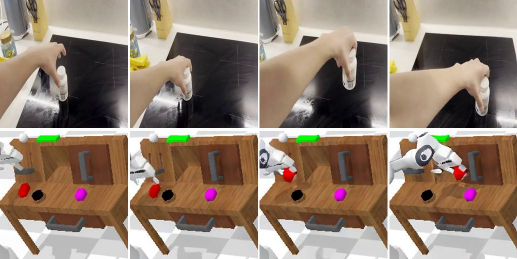}
        & \includegraphics[width=\qatappimgw]{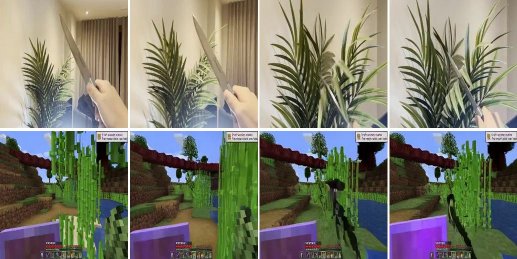}
        & \includegraphics[width=\qatappimgw]{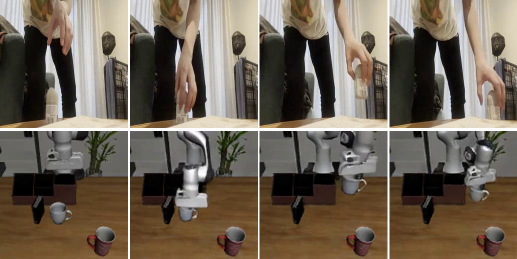} \\[4pt]
    \end{tabular}
    \caption{\textbf{Additional action-transfer results.}  Latent actions extracted from the demonstration frames (top) condition the video world model generations (bottom). \mname{} reproduces the demonstrated actions across various settings. Best seen as videos in the supplementary material which also includes more examples.}
    \label{fig:appendix-action-transfer}
\end{figure*}

\subsection{Whitening Control}
\label{appendix:whitening_control}
Only the \mname{} embeddings are whitened before entering the world model (Appendix~\ref{appendix:implementation_details}), while the baselines in \cref{tab:main_video_quality} use their latent actions unchanged. To check that the whitening is not what drives the comparison, we train a second DreamDojo$^\dagger$ world model whose latent actions pass through the same whitening rule, with statistics computed on its own embeddings of the same training clips, and keep everything else identical (backbone, recipe, data, seed, and sampler). \cref{tab:whitening_control} evaluates it on the same clips and source--target pairs as \cref{tab:main_video_quality}. Whitening changes DreamDojo's results only slightly (by at most 0.23\,dB in Self PSNR and 0.023 in Transfer Dir). \mname{} remains ahead of the whitened DreamDojo in Self PSNR on LIBERO and RLBench and in Transfer Dir on LIBERO, is level with it in Self PSNR on SpatialVID, and keeps a positive but not statistically significant lead in Transfer Dir on SpatialVID. The whitening therefore does not explain the advantage of \mname{}.

\begin{table}[h]
\centering
\caption{\textbf{Whitening control.} DreamDojo$^\dagger$ world models trained with their native latent actions (as in \cref{tab:main_video_quality}) and with the same whitening rule as \mname{} (statistics computed on its own embeddings), everything else identical, evaluated on the same clips and source--target pairs as \cref{tab:main_video_quality} (LIBERO and RLBench: 40; SpatialVID: 20). RLBench transfer and nuScenes were not evaluated for the whitened model (--). $\Delta$: ours minus the whitened DreamDojo, tested with a paired $t$-test over clips (Self PSNR) and a two-sided Wilcoxon signed-rank test over source--target pairs (Transfer Dir); grey: not significant at the 0.05 level. Single training seed.}
\label{tab:whitening_control}
\setlength{\tabcolsep}{4pt}
\renewcommand{\arraystretch}{1.12}
\resizebox{\linewidth}{!}{%
\begin{tabular}{@{}lcccccc@{}}
\toprule
 & \multicolumn{2}{c}{LIBERO$^{\circ}$} & \multicolumn{2}{c}{RLBench} & \multicolumn{2}{c}{SpatialVID$^{\circ}$} \\
\cmidrule(lr){2-3} \cmidrule(lr){4-5} \cmidrule(lr){6-7}
\textbf{Action representation} & {\small Self PSNR$\uparrow$} & {\small Transfer Dir$\uparrow$} & {\small Self PSNR$\uparrow$} & {\small Transfer Dir$\uparrow$} & {\small Self PSNR$\uparrow$} & {\small Transfer Dir$\uparrow$} \\
\midrule
DreamDojo$^\dagger$, native input (\cref{tab:main_video_quality}) & 19.95 & 0.803 & 18.08 & 0.572 & 16.40 & 0.480 \\
DreamDojo$^\dagger$, same whitening as ours & 19.93 & 0.826 & 18.00 & -- & 16.17 & 0.487 \\
\midrule
\rowcolor{oursblue}
\textbf{\mname} (ours) & 20.32 & 0.917 & 18.63 & 0.686 & 16.10 & 0.608 \\
\multicolumn{1}{@{}r}{\footnotesize $\Delta$ vs.\ whitened DreamDojo} & \footnotesize \textcolor{deltapos}{+0.40} & \footnotesize \textcolor{deltapos}{+0.091} & \footnotesize \textcolor{deltapos}{+0.63} & \footnotesize -- & \footnotesize \textcolor{gray}{$-$0.07} & \footnotesize \textcolor{gray}{+0.121} \\
\multicolumn{1}{@{}r}{\footnotesize test statistic} & \footnotesize $t{=}2.42$ & \footnotesize $p{<}10^{-5}$ & \footnotesize $t{=}3.46$ & \footnotesize -- & \footnotesize $t{=}{-}0.31$ & \footnotesize $p{=}0.31$ \\
\bottomrule
\end{tabular}}
\end{table}

\subsection{Multimodal Embedding Backbone}
\label{appendix:mllm_variant}
The submitted version of this work extracted actions from a multimodal embedding model, Qwen3-VL-Embedding-8B~\cite{li2026qwen3embedding}, instead of V-JEPA~2. To compare the two backbones under identical conditions, we apply the \mname{} recipe unchanged to Qwen3-VL-Embedding-8B: the same 168K training pairs, augmentation, positives and hard negatives, loss, optimizer, schedule and 10{,}000 steps (Appendix~\ref{appendix:contrastive_learning}). Only the backbone, its input format and its LoRA targets differ. A frame pair is resized and center-cropped to $256{\times}256$ exactly as for V-JEPA~2 and given to the model as two images after the instruction \emph{``Describe what action is happening in this scene as an embedding.''}; the embedding is the model's own last-token hidden state (4096-d), mapped to 640 dimensions by a fixed random linear layer and $\ell_2$-normalized, as for \mname{}. LoRA adapters with the same rank 16, $\alpha{=}32$ and dropout 0.05 as \mname{} are attached to the query, key, value and MLP projections of the language model (41.6M trainable parameters; with gradient checkpointing, training fits on one RTX 4090). Action expressiveness and scene leakage are evaluated with exactly the protocols, clips, splits and seeds of \cref{tab:action_probing,tab:semantic_leakage}.

\paragraph{Results.} The contrastive task itself is learned: the anchor-to-positive retrieval accuracy among the five in-batch candidates (chance 0.20) rises from 0.72 over the first 500 steps to 0.96 over the last 2{,}000 (0.98 for \mname{} over the same steps). The resulting representation, however, carries much less action information than \mname{} (\cref{tab:mllm_variant}, left): its average probe $R^2$ is 0.142 against 0.490, it is lower on all 13 datasets, and it is also below the frozen V-JEPA~2 embedding (0.347). It lets slightly less of the scene be reconstructed than \mname{} (\cref{tab:mllm_variant}, right; average held-out LPIPS 0.315 vs.\ 0.304), as expected for a representation that carries less information overall, which is why scene leakage has to be read together with action expressiveness: \mname{} gains 0.35 in average $R^2$ for 0.011 in average LPIPS. These results motivated the switch to V-JEPA~2 in the final version; the results reported for the multimodal backbone in the submitted version are superseded by this re-evaluation.

\begin{table}[h]
\centering
\caption{\textbf{The \mname{} recipe on a multimodal embedding backbone.} Left: linear probing of action information (test $R^2$; Procgen: macro-F1; mean over 3 probe seeds) with the protocol, clips, splits and seeds of \cref{tab:action_probing}. Right: scene leakage (held-out LPIPS$\uparrow$, higher = less scene information recoverable; mean over 4 decoder seeds) with the decoders and held-out scenes of \cref{tab:semantic_leakage}. V-JEPA~2: the frozen backbone of \mname{}; \mname{}: main-text results; Qwen recipe: the same recipe (168K training pairs, augmentation, loss, optimizer, 10{,}000 steps) on Qwen3-VL-Embedding-8B (Appendix~\ref{appendix:mllm_variant}). $\Delta$: Qwen recipe minus \mname{}; green: better for the Qwen recipe (higher $R^2$, less leakage). Avg.: probe over the 12 continuous-action datasets, leakage over all 13. $^{\circ}$Videos seen during contrastive training (the same for both backbones).}
\label{tab:mllm_variant}
\small
\setlength{\tabcolsep}{4pt}
\renewcommand{\arraystretch}{1.1}
\begin{tabular}{@{}lc>{\columncolor{oursblue}}cccc>{\columncolor{oursblue}}ccc@{}}
\toprule
 & \multicolumn{4}{c}{\textit{Linear probe} $R^2\uparrow$} & \multicolumn{4}{c}{\textit{Scene leakage}: held-out LPIPS$\uparrow$} \\
\cmidrule(lr){2-5} \cmidrule(lr){6-9}
\textbf{Dataset} & V-JEPA~2 & \mname{} & Qwen recipe & $\Delta$ & V-JEPA~2 & \mname{} & Qwen recipe & $\Delta$ \\
\midrule
LIBERO$^{\circ}$ & 0.718 & 0.634 & 0.443 & \footnotesize \textcolor{deltaneg}{$-$0.19} & 0.257 & 0.314 & 0.323 & \footnotesize \textcolor{deltapos}{+0.009} \\
DROID$^{\circ}$ & $-$0.131 & 0.203 & $-$0.085 & \footnotesize \textcolor{deltaneg}{$-$0.29} & 0.289 & 0.328 & 0.343 & \footnotesize \textcolor{deltapos}{+0.015} \\
RLBench & 0.556 & 0.566 & 0.426 & \footnotesize \textcolor{deltaneg}{$-$0.14} & 0.216 & 0.269 & 0.266 & \footnotesize \textcolor{deltaneg}{$-$0.003} \\
RoboCasa & $-$0.368 & 0.322 & 0.034 & \footnotesize \textcolor{deltaneg}{$-$0.29} & 0.480 & 0.411 & 0.433 & \footnotesize \textcolor{deltapos}{+0.022} \\
DROID-IPRL & 0.186 & 0.279 & $-$0.035 & \footnotesize \textcolor{deltaneg}{$-$0.31} & 0.247 & 0.322 & 0.337 & \footnotesize \textcolor{deltapos}{+0.015} \\
MetaWorld & 0.144 & 0.277 & $-$0.049 & \footnotesize \textcolor{deltaneg}{$-$0.33} & 0.066 & 0.105 & 0.125 & \footnotesize \textcolor{deltapos}{+0.020} \\
MUTEX & 0.527 & 0.566 & 0.229 & \footnotesize \textcolor{deltaneg}{$-$0.34} & 0.130 & 0.177 & 0.197 & \footnotesize \textcolor{deltapos}{+0.020} \\
Robosuite & 0.448 & 0.695 & 0.175 & \footnotesize \textcolor{deltaneg}{$-$0.52} & 0.282 & 0.317 & 0.331 & \footnotesize \textcolor{deltapos}{+0.014} \\
RoboDesk & 0.520 & 0.581 & 0.301 & \footnotesize \textcolor{deltaneg}{$-$0.28} & 0.069 & 0.100 & 0.101 & \footnotesize \textcolor{deltapos}{+0.001} \\
SpatialVID$^{\circ}$ & 0.301 & 0.560 & 0.263 & \footnotesize \textcolor{deltaneg}{$-$0.30} & 0.402 & 0.407 & 0.418 & \footnotesize \textcolor{deltapos}{+0.011} \\
ARKitScenes & 0.665 & 0.588 & 0.104 & \footnotesize \textcolor{deltaneg}{$-$0.48} & 0.384 & 0.411 & 0.424 & \footnotesize \textcolor{deltapos}{+0.014} \\
RE10K & 0.598 & 0.609 & $-$0.102 & \footnotesize \textcolor{deltaneg}{$-$0.71} & 0.442 & 0.436 & 0.418 & \footnotesize \textcolor{deltaneg}{$-$0.018} \\
Procgen (F1) & 0.145 & 0.200 & 0.126 & \footnotesize \textcolor{deltaneg}{$-$0.07} & 0.311 & 0.352 & 0.376 & \footnotesize \textcolor{deltapos}{+0.025} \\
\midrule
Avg. & 0.347 & 0.490 & 0.142 & \footnotesize \textcolor{deltaneg}{$-$0.35} & 0.275 & 0.304 & 0.315 & \footnotesize \textcolor{deltapos}{+0.011} \\
Avg.\ SSIM$\downarrow$ &  &  &  &  & 0.478 & 0.391 & 0.379 & \footnotesize \textcolor{deltapos}{$-$0.012} \\
Avg.\ L1$\uparrow$ &  &  &  &  & 0.200 & 0.255 & 0.266 & \footnotesize \textcolor{deltapos}{+0.010} \\
\bottomrule
\end{tabular}
\end{table}

\section{Experiment Details}
\subsection{Implementation Details}
\label{appendix:implementation_details}
\paragraph{Action encoder.} We use the encoder of V-JEPA~2 ViT-L (\texttt{vjepa2-vitl-fpc64-256}; 24 layers, hidden size 1024, $16{\times}16$ patches, tubelet size 2)~\cite{vjepa2}. Frames are resized to a short side of 292, center-cropped to $256{\times}256$, and normalized with ImageNet statistics using the official V-JEPA~2 preprocessing. A frame pair is stacked into a two-frame clip, which produces a single temporal slice of $16{\times}16{=}256$ tokens; the V-JEPA~2 predictor is not used. The output tokens are average-pooled to a 1024-dimensional vector, projected to 640 dimensions by a linear map that is randomly initialized (seed 0) and kept fixed, and $\ell_2$-normalized.

\paragraph{Low-rank adaptation.} We attach LoRA adapters~\cite{hu2022lora} with rank $16$, $\alpha{=}32$, and dropout $0.05$ to the query, key, value, and output projections of the attention and to both MLP layers in all 24 encoder blocks, as well as to the patch-embedding convolution. The pretrained weights are frozen and kept in bf16. This amounts to 7.1M trainable parameters, about 2.3\% of the 300M encoder parameters.

\paragraph{World model.} We use the AdaWorld world model~\cite{gao2025adaworld} as the backbone for all action representations. We attach LoRA adapters with rank $128$ and $\alpha{=}128$ to the query, key, value, and output projections of the denoiser attention, and train the label-embedding MLP in full, which amounts to 61.5M trainable parameters for \mname{} and 55.7M for the 32-dimensional latent-action baselines. We use AdamW with a learning rate of $5{\times}10^{-5}$, 1k linear warm-up steps followed by a constant schedule, a batch of 4 clips of 7 frames, 20k training steps, bf16 precision, gradient clipping at 0.3, no EMA, and seed 32. All world models are trained on the same 19{,}946 clips (25 frames, $256{\times}256$) from 9 datasets. Before entering the world model, the \mname{} embeddings are whitened by a fixed, parameter-free covariance equalization. Each dimension is first standardized; the covariance $C = Q\,\mathrm{diag}(e)\,Q^{\top}$ of the standardized embeddings $z$ is then equalized as $y = z\,Q\,\mathrm{diag}(g)\,Q^{\top}$ with gains $g_i = s/\sqrt{e_i + r}$, where the ridge $r \ge 0$ is the smallest value for which no gain exceeds 2, and $s$ keeps the total variance equal to the dimensionality, so that no direction is removed. All statistics are computed once on the 478{,}704 frame pairs of the world-model training clips. The latent actions of AdaWorld and DreamDojo enter the world model unchanged; Appendix~\ref{appendix:whitening_control} reports a DreamDojo world model trained with the same whitening. Sampling uses the Euler EDM sampler with 25 steps, a guidance scale of 1.1, and 6 context frames, rolled out autoregressively for 19 frames. Each world model takes about 5 GPU-hours on a single RTX 4090.

\paragraph{Compute.} All experiments are run on single 24\,GB GPUs (NVIDIA RTX 4090 or RTX A5500). Encoder adaptation takes about 1.5 GPU-hours on an RTX 4090.

\subsection{Contrastive Learning Details}
\label{appendix:contrastive_learning}
\paragraph{Training data.} No action labels are used in this stage. We pre-sample 168K training samples from nine video sources with the following sampling weights: VPT Minecraft~\cite{vpt} 0.25, CALVIN~\cite{mees2022calvin} 0.15, SpatialVID~\cite{spatialvid} 0.15, AgiBot World~\cite{agibot} 0.10, EgoDex~\cite{egodex} 0.10, Something-Something v2~\cite{sthv2} 0.10, and LIBERO~\cite{libero}, DROID~\cite{droid}, and Atari~\cite{atari} 0.05 each.

\paragraph{Anchor.} The anchor is a frame pair $(o_t, o_{t+k})$ from one video, with $k$ sampled uniformly from $\{1,2,3\}$. Both frames receive the same color jitter (brightness, contrast, and saturation $\pm 0.3$, hue $\pm 0.05$).

\paragraph{Positive: flow-guided scene swapping.} We first decide whether the camera is static by the mean optical-flow magnitude of the background: if it is at most 0.5 pixels, the camera is treated as static. In this case, we compute the moving core region as the pixels whose flow magnitude exceeds $\max(1\,\text{px},\, 0.35 \times q_{99})$, where $q_{99}$ is the 99th percentile of the flow magnitude, without dilation or feathering. We cut out the original pixels of this region from both frames and paste them onto the same background frame, taken from a different scene of the same dataset, translated in both frames by a common random integer offset of at least 16 pixels. The moving pixels are therefore preserved exactly, while the scene and the absolute location of the motion are replaced. A color jitter shared between the two frames is applied afterwards. When the camera moves (e.g., hand-held, egocentric, or driving video), the background motion is itself part of the action and cannot be replaced; the positive then falls back to the anchor pair under a different color-jitter draw. During training, about 43\% of samples use the scene-swapping path (mean over the training log): it dominates on static-camera datasets (LIBERO 99\%, DROID 88\%, Atari 75\%, AgiBot World 62\%), whereas hand-held and ego-motion datasets mostly fall back. Optical flow is precomputed with RAFT-large~\cite{teed2020raft} (12 iterations, $256{\times}256$ input).

\paragraph{Hard negative.} The hard negative is a frame pair from a different, non-overlapping time window of the same video as the anchor. It shares the scene with the anchor but generally corresponds to a different action.

\paragraph{Objective and optimization.} We use the symmetric InfoNCE loss of \cref{sec:method_contrastive} with temperature $\tau{=}0.07$, over a candidate bank formed by all anchors, positives, and hard negatives in the batch. We train on a single RTX 4090 with 2 samples per step, so that each query is contrasted against one positive and four negatives: its own same-video hard negative and the anchor, positive, and hard negative of the other sample; we accumulate gradients over 8 steps for an effective batch size of 16 (gradient accumulation does not enlarge the negative set, and no memory queue is used). We use AdamW with learning rate $10^{-4}$, 500 warm-up steps followed by cosine decay to zero, no weight decay, and seed 0, for 10{,}000 updates (about 160K samples, roughly one pass over the training set). At the end of training, the positive is retrieved for every query (chance: 0.2), and the mean cosine similarity is 0.88 to the positive and 0.215 to the hard negative, compared with about 0.94 for both at initialization.

\subsection{Action Expressiveness Details}
\label{appendix:action_expressiveness}
  For each embedding family, we train an independent linear probe $W \in \mathbb{R}^{D \times K}$ (zero-initialized weight and bias) to map the per-pair embedding $z_t \in \mathbb{R}^{D}$ (for \mname{}, $z_t = \alpha_t$ from \cref{sec:method_encoder}) to the ground-truth action $a_t$, where $D$ is the embedding dimensionality and $K$ is the action dimensionality for continuous datasets or the number of classes for the discrete datasets. Inputs are $z$-scored using mean and standard deviation estimated from a $20{,}000$-sample subset of the training split; for continuous datasets, the targets are $z$-scored with the same procedure, and for the discrete datasets, we apply softmax cross-entropy directly. We train for $50$ epochs with AdamW (learning rate $10^{-2}$, weight decay $10^{-3}$) and batch size $4{,}096$, without early stopping, and report the last epoch averaged over 3 probe seeds on a fixed test split. The probe is deliberately kept linear so that performance differences reflect representation quality rather than probe capacity.

\subsection{Scene Invariance Details}
\label{appendix:scene_invariance}
  To measure how much pixel-level appearance can be reconstructed from the per-pair embedding $z_t \in \mathbb{R}^{D}$, we train a small convolutional decoder $g_\phi : \mathbb{R}^{D} \to \mathbb{R}^{3 \times 128 \times 128}$ to recover the second frame
  $o_{t+1}$ of the pair. The decoder consists of a single linear projection from $z_t$ to a $512{\times}4{\times}4$ feature map, followed by five upsampling blocks ($4{\to}8{\to}16{\to}32{\to}64{\to}128$); each block applies a $4{\times}4$ stride-$2$ ConvTranspose2d, GroupNorm, GELU, a $3{\times}3$ Conv2d, GroupNorm, and GELU. A final $3{\times}3$ convolution with $\tanh$ produces the RGB output. We train with a combined L1 + LPIPS loss with $\lambda_{\text{LPIPS}}{=}0.5$, using AdamW (learning rate $10^{-4}$, weight decay $10^{-4}$) with cosine annealing to zero and batch size $64$, for $10$ epochs per (embedding, dataset) pair. We report the last-epoch metrics averaged over 4 seeds, each with its own random split that holds out 20\% of the scenes for evaluation.

\subsection{Real-Action Fine-Tuning Details}
\label{appendix:real_action_control}
\paragraph{Domains and data.} DMLab and Minecraft use the released datasets of TECO~\cite{teco} (DMLab actions: look left, look right, forward; Minecraft actions: forward with jump, camera left $20^\circ$, camera right $20^\circ$). Habitat uses frames rendered with Habitat~\cite{habitat} in the HM3D example scenes~\cite{hm3d} (move forward, turn left, turn right), with fine-tuning clips from one scene and test cases from two other scenes. nuScenes uses the front camera, with the ego motion (forward and lateral displacement in meters) as a continuous action. Each discrete-action domain has 300 labeled fine-tuning clips (100 per action) and 300 test cases; nuScenes has 100 fine-tuning clips from 4 driving logs and 60 test cases from 12 other driving logs.

\paragraph{Action interface.} For the discrete-action domains, the interface is an action book with one row per action. Each row is initialized at the mean of the representation's latent actions, after the world model's input transform, over the 100 labeled clips of that action. The \emph{0 updates} column of \cref{tab:downstream_finetune} uses this initialization without any gradient update, so it already uses the labeled clips. For nuScenes, the interface is a two-layer MLP (linear, SiLU, linear) from the two-dimensional ego motion to the latent-action space, fitted before fine-tuning by 3{,}000 SGD steps (learning rate 0.01, batch 10, mean-squared error) to regress the transformed latent actions of the labeled clips.

\paragraph{Two recipes.} DMLab and Minecraft use a \emph{standardized} interface: the book stores $u$, and the condition is $\mu + \sigma \odot u$, where $\mu$ and $\sigma$ are the per-dimension statistics of each representation's own world-model training features. The book is trained with a learning rate of $2.42{\times}10^{-3}$, so that every representation receives the same step size relative to the scale of its features. Habitat and nuScenes use the \emph{native} interface: the book rows or the MLP output the condition directly and are trained with a learning rate of $5{\times}10^{-5}$.

\paragraph{Optimization.} In both recipes, the whole denoiser (1.56B parameters) is fine-tuned with a learning rate of $5{\times}10^{-6}$ for 800 AdamW updates (weight decay 0.01) with a global batch of 32 clips, 100 linear warm-up updates, gradient clipping at 0.3, an EMA of the denoiser weights (decay 0.9999), bf16 mixed precision, and seed 32. The 200- and 800-update checkpoints come from the same run, and each cell of \cref{tab:downstream_finetune} is a single fine-tuning run.

\paragraph{Evaluation.} Each test case has six real context frames, and the next frame is generated under every candidate action with the Euler EDM sampler (5 steps, guidance scale 1.05), using the same noise for all actions and all representations. On the discrete-action domains, Follow uses an inverse-dynamics classifier trained only on real frames, which receives the last real context frame and the generated frame. On nuScenes, Follow is the per-case Spearman correlation between five commanded forward displacements and the forward displacement that a regressor trained only on real frames reads from the last real context frame and the generated frame. Control is the PSNR under the commanded action minus the mean PSNR under the two other actions (on nuScenes, under four actions drawn from the training clips). $\Delta$ is tested with a paired bootstrap over test episodes (driving logs on nuScenes; 20{,}000 resamples), which covers the sampling of test cases but not the variance between training runs.

\section{Licenses of Existing Assets}
\label{appendix:licenses}
\cref{tab:licenses} lists the license or terms of use of every dataset, model, and code base that we use, as stated by its source. Several datasets are restricted to non-commercial use (AgiBot World, EgoDex, SpatialVID, ARKitScenes, nuScenes, HM3D, and RLBench); we use them for non-commercial research only. We do not redistribute any data: our release contains only clip lists, splits, and preparation scripts that read each dataset from its original source.

\begin{table}[h]
\centering
\caption{\textbf{Licenses of existing assets}, as stated by each source.}
\label{tab:licenses}
\small
\renewcommand{\arraystretch}{1.08}
\begin{tabular}{@{}p{0.31\linewidth}p{0.20\linewidth}p{0.43\linewidth}@{}}
\toprule
\textbf{Asset} & \textbf{Used for} & \textbf{License or terms} \\
\midrule
VPT Minecraft~\cite{vpt} & training & MIT (repository; no separate license for the recordings) \\
CALVIN~\cite{mees2022calvin} & training & MIT \\
AgiBot World~\cite{agibot} & training & CC BY-NC-SA 4.0 \\
EgoDex~\cite{egodex} & training & CC BY-NC-ND \\
Something-Something v2~\cite{sthv2} & training & Qualcomm license agreement \\
Atari~\cite{atari} & training & none stated \\
LIBERO~\cite{libero} & training, evaluation & CC BY 4.0 \\
DROID~\cite{droid} & training, evaluation & CC BY 4.0 \\
SpatialVID~\cite{spatialvid} & training, evaluation & CC BY-NC-SA 4.0 \\
RLBench~\cite{james2019rlbench} (PerAct demonstrations) & evaluation & Imperial College London non-commercial research license; none stated for the pre-generated demonstrations \\
DROID-IPRL~\cite{droid} & evaluation & CC BY 4.0 \\
RoboCasa~\cite{robocasa} & evaluation & CC BY 4.0 \\
MetaWorld~\cite{metaworld} (LeRobot release) & evaluation & Apache-2.0 \\
MUTEX~\cite{mutex} & evaluation & Apache-2.0 \\
Robosuite~\cite{robosuite}, RoboDesk~\cite{robodesk} (VP$^2$ data) & evaluation & MIT \\
ARKitScenes~\cite{arkitscenes} & evaluation & Apple ARKitScenes license (non-commercial terms) \\
RealEstate10K~\cite{realestate} & evaluation & camera files CC BY 4.0; frames from YouTube, no license stated \\
Procgen~\cite{cobbe2019procgen} & evaluation & MIT \\
nuScenes~\cite{nuscenes} & evaluation, fine-tuning & CC BY-NC-SA 4.0 and the nuScenes Terms of Use \\
DMLab~\cite{dmlab}, Minecraft (TECO data~\cite{teco}) & fine-tuning & none stated \\
HM3D example scenes~\cite{hm3d} & fine-tuning & Matterport academic, non-commercial license \\
\midrule
V-JEPA~2 ViT-L~\cite{vjepa2} & encoder backbone & MIT \\
AdaWorld~\cite{gao2025adaworld} & world model, baseline & Apache-2.0 (weights fine-tuned from Stable Video Diffusion; Stability AI terms apply) \\
DreamDojo~\cite{gao2026dreamdojo} & baseline & Apache-2.0 (code) \\
RAFT~\cite{teed2020raft} (torchvision) & positive construction & BSD-3-Clause \\
\bottomrule
\end{tabular}
\end{table}

\section{Limitations}
\label{appendix:limitations}
\mname{} demonstrates that action representations can be extracted from a pretrained foundation model and adapted for action-conditioned video world models without explicit action labels. However, several limitations remain. First, the method inherits both the strengths and weaknesses of the underlying foundation model. Although large-scale video pretraining provides broad spatio-temporal priors that improve transfer, the resulting action space may still reflect biases in the pretraining distribution, and its performance may degrade for domains with visual dynamics, embodiments, or interaction patterns that are poorly represented in that data. Second, \mname{} relies on positive pairs that preserve the underlying action while changing the scene. Flow-guided scene swapping is only applicable to static cameras; for moving cameras we fall back to photometric augmentation, which provides a weaker scene-invariance signal. Optical-flow errors, thin or occluded moving parts, and interactions whose effect extends beyond the moving region may also make the swapped positive imperfect. Third, the contrastive stage uses a small batch with few negatives per query and saturates before the end of training, so larger batches, negative queues, or harder negatives may further improve the representation. Fourth, while contrastive adaptation substantially reduces scene leakage, it does not provide a formal guarantee that the learned representation is fully disentangled from scene content or future-state information.

In addition, real-action control still requires a small amount of action-labeled data to align environment-specific actions with the learned latent-action space. While this requirement is substantially weaker than training fully supervised action-conditioned models, it means that \mname{} is not entirely label-free for direct control. Finally, our experiments cover robot manipulation, camera navigation, game, and driving data, but they remain offline evaluations. Further work is needed to evaluate robustness in long-horizon closed-loop interaction, real-world robotics, safety-critical settings, and domains involving complex multi-agent behavior.

\section{Broader Impact}
\label{appendix:broader_impact}
This work aims to reduce the dependence of video world models on expensive and fragmented action annotations. By extracting action representations from unlabeled video, \mname{} could make it easier to train controllable simulators across domains where explicit action labels are difficult to obtain, including robotics, embodied AI, game environments, and interactive content creation. In robotics, such representations may support more scalable learning from observational video and lower the annotation burden for adapting world models to new embodiments. In gaming and simulation, they may enable more reusable action interfaces across visually diverse environments.

At the same time, more capable action-conditioned video world models raise several risks. If deployed without safeguards, they could be used to generate misleading interactive simulations or synthetic videos that appear physically plausible but do not faithfully reflect real-world dynamics. In robotics or embodied decision-making, inaccurate predictions could lead to unsafe behavior if treated as reliable simulators without appropriate validation. Because \mname{} builds on a foundation model pretrained on large-scale web video, it may also inherit biases from that data, potentially leading to uneven performance across environments, objects, cultures, or human activities. Moreover, models that learn action structure from unlabeled video may amplify privacy concerns if trained on videos containing people or sensitive environments.

We therefore view \mname{} primarily as a research step toward scalable action representation learning rather than a ready-to-deploy simulator for safety-critical decision-making. Responsible use should include careful dataset curation, evaluation for domain-specific failure modes, transparency about synthetic generation, and validation before deployment in real-world control systems. Releasing code, models, and evaluation protocols can help the community better audit the learned representations, reproduce results, and study both the benefits and risks of annotation-free action-conditioned world modeling.

\newpage
\section*{NeurIPS Paper Checklist}

\begin{enumerate}

\item {\bf Claims}
    \item[] Question: Do the main claims made in the abstract and introduction accurately reflect the paper's contributions and scope?
    \item[] Answer: \answerYes{}
    \item[] Justification: The abstract and \cref{sec:method} state the three claims of the paper---extracting action representations from a pretrained foundation model (V-JEPA~2), a label-free contrastive adaptation that disentangles action from scene without a reconstruction bottleneck, and a unified evaluation across robot, navigation, game, and driving benchmarks---which are matched by the method in \cref{sec:method} and the experiments in \cref{exp:scene_invariance} and the surrounding subsections.

\item {\bf Limitations}
    \item[] Question: Does the paper discuss the limitations of the work performed by the authors?
    \item[] Answer: \answerYes{}
    \item[] Justification: The limitations are discussed in Appendix~\ref{appendix:limitations}, including dependence on the pretrained foundation model's priors, the static-camera requirement of flow-guided scene swapping, the small contrastive batch, imperfect action disentanglement, limited action supervision requirements, and the need for broader real-world evaluation.

\item {\bf Theory assumptions and proofs}
    \item[] Question: For each theoretical result, does the paper provide the full set of assumptions and a complete (and correct) proof?
    \item[] Answer: \answerNA{}
    \item[] Justification: The paper presents an empirical method and evaluation; it contains no formal theorems or proofs.

    \item {\bf Experimental result reproducibility}
    \item[] Question: Does the paper fully disclose all the information needed to reproduce the main experimental results of the paper to the extent that it affects the main claims and/or conclusions of the paper (regardless of whether the code and data are provided or not)?
    \item[] Answer: \answerYes{}
    \item[] Justification: \cref{sec:method} fully specifies the action encoder, contrastive objective, and action-conditioned video world model, and Appendix~\ref{appendix:implementation_details} and~\ref{appendix:contrastive_learning} report the encoder (V-JEPA~2 ViT-L), the LoRA configuration, the positive/negative construction, the contrastive training data and hyperparameters, the world-model backbone (AdaWorld), and the evaluation protocols used for all reported numbers.

\item {\bf Open access to data and code}
    \item[] Question: Does the paper provide open access to the data and code, with sufficient instructions to faithfully reproduce the main experimental results, as described in supplemental material?
    \item[] Answer: \answerYes{}
    \item[] Justification: The code, the trained checkpoints, and the evaluation protocol of every table are released at \url{https://github.com/runjiali-rl/onlyact} (linked from the project page on the first page), with the exact software environments and one script per table. All datasets and the underlying pretrained models (V-JEPA~2, AdaWorld, DreamDojo) are publicly available; we do not redistribute any frames, and the repository provides the clip lists, splits, and preparation scripts of each dataset.

\item {\bf Experimental setting/details}
    \item[] Question: Does the paper specify all the training and test details (e.g., data splits, hyperparameters, how they were chosen, type of optimizer) necessary to understand the results?
    \item[] Answer: \answerYes{}
    \item[] Justification: \cref{sec:experiments} describes the datasets, splits (held-out scenes; held-out demonstrations on LIBERO and RLBench), baselines, and metrics, and Appendix~\ref{appendix:implementation_details}--\ref{appendix:scene_invariance} report the encoder and world-model hyperparameters, the contrastive training data and optimization, and the probe and decoder protocols.

\item {\bf Experiment statistical significance}
    \item[] Question: Does the paper report error bars suitably and correctly defined or other appropriate information about the statistical significance of the experiments?
    \item[] Answer: \answerYes{}
    \item[] Justification: Probing and scene-leakage results are averaged over 3 and 4 seeds, respectively, with the probe standard deviation reported in \cref{tab:action_probing}; in \cref{tab:downstream_finetune}, differences whose 95\% paired bootstrap interval contains zero are marked. World-model training runs (\cref{tab:main_video_quality}) and fine-tuning runs (\cref{tab:downstream_finetune}) are single runs due to compute constraints, so the bootstrap intervals of \cref{tab:downstream_finetune} cover the sampling of test cases but not the variance between training runs.

\item {\bf Experiments compute resources}
    \item[] Question: For each experiment, does the paper provide sufficient information on the computer resources (type of compute workers, memory, time of execution) needed to reproduce the experiments?
    \item[] Answer: \answerYes{}
    \item[] Justification: Appendix~\ref{appendix:implementation_details} reports that all experiments run on single 24\,GB GPUs (RTX 4090 / RTX A5500): encoder adaptation takes about 1.5 GPU-hours and each world model about 5 GPU-hours.

\item {\bf Code of ethics}
    \item[] Question: Does the research conducted in the paper conform, in every respect, with the NeurIPS Code of Ethics \url{https://neurips.cc/public/EthicsGuidelines}?
    \item[] Answer: \answerYes{}
    \item[] Justification: The work uses publicly released datasets and pretrained models within their licensed terms, involves no human subjects, and does not pose obvious risks beyond those of the underlying video generation backbone.

\item {\bf Broader impacts}
    \item[] Question: Does the paper discuss both potential positive societal impacts and negative societal impacts of the work performed?
    \item[] Answer: \answerYes{}
    \item[] Justification: The broader impacts are discussed in Appendix~\ref{appendix:broader_impact}, including the opportunities enabled by scalable action representation learning from unlabeled video, as well as the associated challenges of reliability, bias, misuse, and responsible deployment.

\item {\bf Safeguards}
    \item[] Question: Does the paper describe safeguards that have been put in place for responsible release of data or models that have a high risk for misuse (e.g., pre-trained language models, image generators, or scraped datasets)?
    \item[] Answer: \answerNA{}
    \item[] Justification: The released models are an action encoder and an action-conditioned video world model trained on public robotics, game, and human video data; their misuse risk is comparable to that of the underlying AdaWorld world model, and they are not general-purpose photorealistic generators.

\item {\bf Licenses for existing assets}
    \item[] Question: Are the creators or original owners of assets (e.g., code, data, models), used in the paper, properly credited and are the license and terms of use explicitly mentioned and properly respected?
    \item[] Answer: \answerYes{}
    \item[] Justification: All datasets and pretrained models (V-JEPA~2, AdaWorld, DreamDojo) used in the paper are cited to their original publications, and Appendix~\ref{appendix:licenses} lists the license or terms of use of every asset as stated by its source. Datasets restricted to non-commercial use are used for non-commercial research only, and no data is redistributed.

\item {\bf New assets}
    \item[] Question: Are new assets introduced in the paper well documented and is the documentation provided alongside the assets?
    \item[] Answer: \answerYes{}
    \item[] Justification: The new assets---the contrastively adapted action encoder and the action-conditioned video world model---are documented in \cref{sec:method} (architecture, training objective) and Appendix~\ref{appendix:implementation_details} (hyperparameters); the code, weights, and documentation (environments, training and evaluation commands, checksums, and licenses) are released at \url{https://github.com/runjiali-rl/onlyact}, with the code under the Apache License 2.0.

\item {\bf Crowdsourcing and research with human subjects}
    \item[] Question: For crowdsourcing experiments and research with human subjects, does the paper include the full text of instructions given to participants and screenshots, if applicable, as well as details about compensation (if any)?
    \item[] Answer: \answerNA{}
    \item[] Justification: The paper does not involve crowdsourcing or research with human subjects.

\item {\bf Institutional review board (IRB) approvals or equivalent for research with human subjects}
    \item[] Question: Does the paper describe potential risks incurred by study participants, whether such risks were disclosed to the subjects, and whether Institutional Review Board (IRB) approvals (or an equivalent approval/review based on the requirements of your country or institution) were obtained?
    \item[] Answer: \answerNA{}
    \item[] Justification: The paper does not involve crowdsourcing or research with human subjects, so no IRB review was required.

\item {\bf Declaration of LLM usage}
    \item[] Question: Does the paper describe the usage of LLMs if it is an important, original, or non-standard component of the core methods in this research? Note that if the LLM is used only for writing, editing, or formatting purposes and does \emph{not} impact the core methodology, scientific rigor, or originality of the research, declaration is not required.
    \item[] Answer: \answerNA{}
    \item[] Justification: LLMs were used only for assistance with writing, editing, and formatting, and did not influence the core methodology, scientific rigor, or originality of the research.

\end{enumerate}

\end{document}